\documentclass{article}
\usepackage[T1]{fontenc}
\usepackage{iclr2027_conference,times}

\iclrfinalcopy

\usepackage{amsmath,amsfonts,bm}

\def\eqref#1{equation~\ref{#1}}

\def\1{\bm{1}}

\DeclareMathAlphabet{\mathsfit}{\encodingdefault}{\sfdefault}{m}{sl}
\SetMathAlphabet{\mathsfit}{bold}{\encodingdefault}{\sfdefault}{bx}{n}

\usepackage{amsmath}
\usepackage{amssymb}
\usepackage{booktabs}
\usepackage{threeparttable}
\usepackage{graphicx}
\usepackage{multirow}
\usepackage{float}
\usepackage{hyperref}
\usepackage{url}
\usepackage{xcolor}
\usepackage{colortbl}
\usepackage{wrapfig}
\usepackage{needspace}
\input{inline_layout}

\newcommand{\method}{S2PD}
\newcommand{\methodfull}{Serial-to-Parallel Diffusion}

\title{S2PD: Serial-to-Parallel Diffusion \\
for Physically and Logically Consistent\\
Video Generation}

\author{
Jeffrey Hu\textsuperscript{1} \quad
Daniel Olmeda Reino\textsuperscript{2} \quad
Ayush Tewari\textsuperscript{1} \\
\textsuperscript{1}University of Cambridge \quad
\textsuperscript{2}Toyota Motor Europe
}

\begin{document}
\setlength{\textfloatsep}{1.25\baselineskip}

\maketitle

  \lhead{Preprint}

\begin{abstract}
Bidirectional video diffusion models denoise entire videos in parallel, yet
when trained on effectively unlimited in-distribution data from procedural
generators, continue to violate physical laws and simple symbolic rules. We
introduce \methodfull{} (\method{}), which performs autoregressive
diffusion at high noise before switching to parallel diffusion at
low noise. The autoregressive phase provides the serial computation needed to
coordinate interdependent events and produce valid state transitions while the
parallel phase jointly refines the entire video and reduces sampling time
relative to fully serial generation.
We implement \method{} with
two architectures: a pixel-space diffusion transformer trained from scratch
and a pretrained video model adapted through LoRA fine-tuning with causal attention.
Across games, physical simulations, and real video, \method{} follows
rules more reliably than matched bidirectional baselines and generates videos
with greater temporal stability and sampling efficiency than other serial methods.
\end{abstract}

\section{Introduction}
\label{sec:introduction}

Bidirectional diffusion transformers such as Wan~\citep{wanteam2025wan} are widely used
for large-scale video generation. These models
denoise all video tokens in parallel. Although these
models achieve high visual fidelity as measured by FVD~\citep{unterthiner2018fvd}
and VBench~\citep{huang2024vbench}, multiple studies document morphing artifacts
and violations of physical laws in their samples~\citep{bansal2026videophy2,motamed2026physics}. We discover a more
puzzling result: bidirectional video diffusion transformers continue to
exhibit these failures even when trained on effectively unlimited
procedurally generated data from relatively simple simulations. Physical
accuracy therefore does not emerge from data scaling alone.

The Serial Scaling Hypothesis~\citep{liu2026serialscaling} offers a theoretical
explanation. Liu et al. prove that diffusion models remain in
\(\mathrm{TC}^0\), a class of computations with bounded sequential depth, even
with arbitrarily many denoising steps. For standard diffusion models, increasing
model depth is the only way to supply the serial computation some problems require.
Videos can contain long chains of causally dependent events. We
hypothesize that bidirectional diffusion models, early in denoising when the
video's overall structure is ambiguous, assign locally plausible but mutually
incompatible events to different space--time regions. Subsequent denoising
sharpens these commitments without reconciling them, leading to abrupt changes
or invalid transitions.

Our solution is to let the model commit to structure serially by generating
blocks of video tokens autoregressively. Concurrent work on the Seriality Gap~\citep{diazchao2026seriality}
shows that autoregressive generation with increasingly small block sizes reduces
errors on colliding ball simulations. We implement two
autoregressive diffusion methods and demonstrate a similar pattern across a broad
range of games and simulations.
We further adapt a large
pretrained bidirectional video model for autoregressive generation through
LoRA fine-tuning~\citep{hu2022lora} on more complex simulations and real video.
On real video of a Rubik's Cube, our models generate realistic hand--cube
interactions while maintaining a plausible cube state over time.

Finally, we introduce \methodfull{} (\method{}). Although autoregressive
diffusion models follow rules much better than bidirectional
diffusion models, they are slow to sample and susceptible to temporal instability and autoregressive
drift~\citep{huang2025selfforcing}. Autoregressive drift occurs when generated content is used as context for
generating more content later. Small prediction errors accumulate, causing sampling and training
distributions to diverge.
Re-noising previously generated blocks can mitigate this drift but it also corrupts
the information available to subsequent blocks, causing flicker.
\method{} divides sampling into two phases, confining the autoregressive
diffusion essential for logical consistency to the high-noise regime, before
refining the full sequence in parallel at low noise.

\begin{figure}[tbp]
  \centering
  \includegraphics[width=\linewidth]{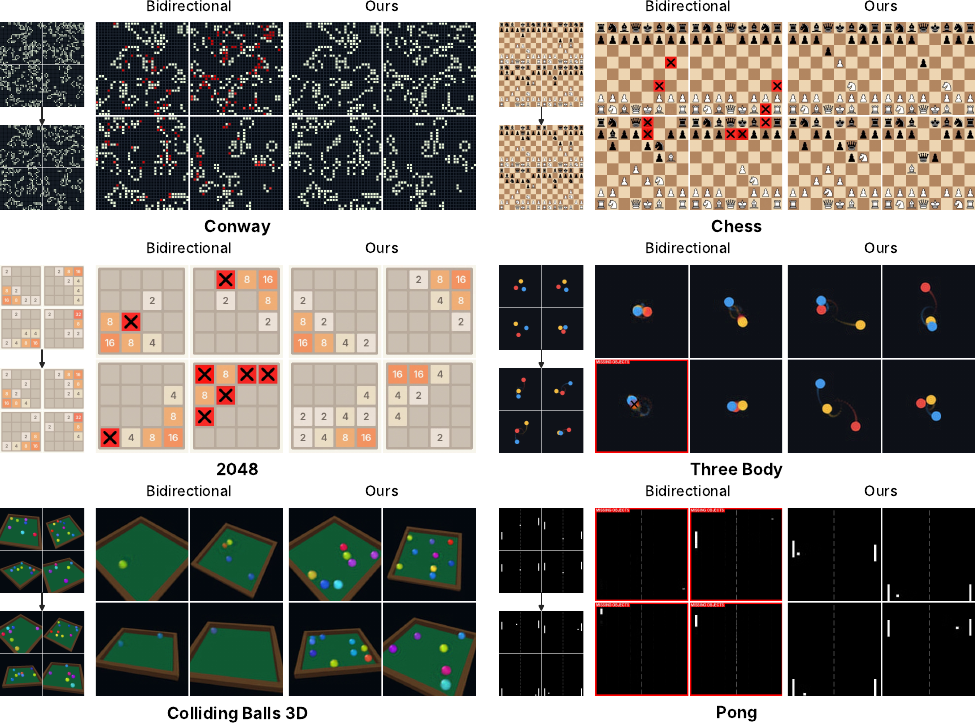}
  \caption{\textbf{Serial outperforms parallel generation.} Our method better preserves game rules, physical dynamics, and object consistency across all datasets. Red overlays localize errors in the generated samples.}
  \label{fig:teaser}
\end{figure}

Our contributions are as follows:
\begin{itemize}
  \item We introduce \method{}, which preserves the physical
  plausibility and logical consistency of serial generation while improving
  sampling efficiency and temporal stability.
  \item We show that bidirectional diffusion violates physical and
  symbolic rules despite effectively unlimited procedurally generated
  in-distribution data, whereas autoregressive diffusion methods
  substantially reduce these errors.
  \item We provide datasets of games and simulations, along with automated
  evaluation metrics for rule violations and physical accuracy.
  We have released all code, datasets, and model weights.
\end{itemize}

\section{Related Work}
\label{sec:related}

\paragraph{Autoregressive generation.}
Video Diffusion Models~\citep{ho2022video} extends image diffusion~\citep{ho2020ddpm} to jointly
denoise all video tokens in parallel. In contrast, autoregressive video models
generate tokens sequentially, using classification for discrete
tokens~\citep{yan2021videogpt} or diffusion for continuous tokens.
Diffusion Forcing~\citep{chen2024diffusionforcing} achieves autoregressive
video diffusion by training a model to denoise independently noised frames.
Self Forcing~\citep{huang2025selfforcing} points out that repeated autoregressive
denoising leads to a training--inference distribution mismatch and addresses
this by also training on generated video.
Block Diffusion~\citep{arriola2025blockdiffusion} introduces a method for
generating chunks of text simultaneously to accelerate language modeling.
We find block-causal diffusion particularly well suited to video modeling since
it can generate whole frames at once.
Flex-Forcing~\citep{ma2026flexforcing} uses block-causal diffusion with
variable block sizes, but they only explore block sizes larger than one
frame and use it as an acceleration technique with
efficiency--quality trade-offs. We explore block sizes smaller than one frame
and study how spatio-temporal autoregression affects logical consistency.

\paragraph{World modeling.}
\citet{oh2015actionconditional} demonstrate action-conditioned video generation
in Atari games.
World models such as Genie 3~\citep{deepmind2025genie3} and
Cosmos 3~\citep{nvidia2026cosmos3} have scaled this to richer environments.
These models are naturally autoregressive to allow users to submit actions
that affect the generated video stream in real time. Thus,
they already support the serial computation necessary for logical consistency.
MIRA~\citep{hu2026mira} demonstrates remarkable physics consistency in the highly
dynamic game of Rocket League. Our work explains the role of serial computation
in achieving such high fidelity and shows that video models can achieve
strong physical and logical consistency without a temporally dense
action-conditioning signal. World models may also benefit from a
parallel refinement stage.

Frontier coding models such as GPT-6 Astra~\citep{openai2026astra} and
Claude Fable 5~\citep{anthropic2026fable5} can also reconstruct scenes
and run simulations through code and software tools.
VDAWorld~\citep{omahony2026vdaworld} and VIGA~\citep{yin2026viga} construct
executable programs to represent scenes explicitly before rendering them.
VLMs supply the required serial computation through text. We achieve it through
autoregressive video generation without relying on human-labeled expert trajectories.

\paragraph{Evaluating physical consistency.}
Numerous benchmarks such as Physics-IQ~\citep{motamed2026physics},
PhyWorldBench~\citep{gu2026phyworldbench}, T2VPhysBench~\citep{guo2025t2vphysbench},
and PhyGround~\citep{lin2026phyground} test whether video models correctly
capture physical laws and dynamics.
Although these benchmarks remain unsaturated, their real-world
setting and generalization demands make it difficult to isolate the causes of
failure. We study simpler games and simulations, where video parsing provides quantitative feedback on rule violations
and physical error, before scaling up to more complex datasets.

\section{Method}
\label{sec:method}

\subsection{Background}
\label{sec:method-background}

We adopt the modern flow-matching~\citep{lipman2023flowmatching} formulation of diffusion.
Let \(x_0\) denote a clean target.
We corrupt \(x_0\) with Gaussian noise \(\epsilon\sim\mathcal{N}(0,I)\)
along a linear path
\begin{equation}
  x_t=(1-t)x_0+t\epsilon,
  \label{eq:flow-path}
\end{equation}
where \(t=0\) denotes clean data and \(t=1\) denotes noise. Following
JiT~\citep{li2026backtobasics}, we use \(x\)-prediction with \(v\)-loss:
the model predicts the clean target \(\widehat x_0=f_\theta(x_t,t)\), while
the loss is computed with mean squared error in velocity space.
We integrate the velocity \(\widehat v_\theta=(x_t-\widehat x_0)/t\)
from \(t=1\) to \(t=0\) during sampling.

For video generation, we patchify the video into \emph{tubelets}: spatial
patches of size \(p_h\times p_w\) spanning \(p_t\) consecutive frames.
Each tubelet becomes one transformer token.
Smaller \(p_t\) provides finer temporal granularity at the cost of increased
token count, which increases training and inference times.
Inspired by JiT's use of large image patches, we send the full
\(3p_t p_h p_w\)-dimensional tubelet through our model. We use a linear
bottleneck before projecting to the transformer width.

\paragraph{Token order and block size.}
When flattening a 3D video token grid into a sequence, we first define a
consistent token order and use 3D RoPE to retain information about
the tokens' spatial and temporal
positions after flattening.
See Section~\ref{sec:token-order} for the token order ablation.
We then group consecutive tokens into \(K\) blocks
\(B^0,\ldots,B^{K-1}\) of block size \(b\).
Tokens within a block share a noise level
during training and are denoised together during inference.
Smaller blocks allow more autoregressive steps during sampling.
When the block size equals the total sequence length, generation becomes equivalent
to full-video bidirectional diffusion.

\begin{figure}[tbp]
  \centering
  \includegraphics[width=\linewidth]{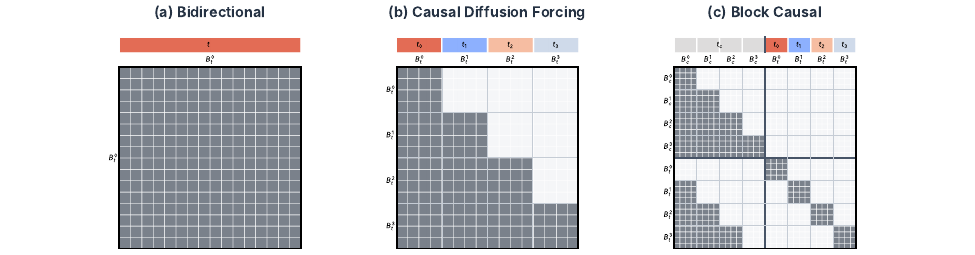}\par
  \vspace{6pt}
  \includegraphics[width=\linewidth]{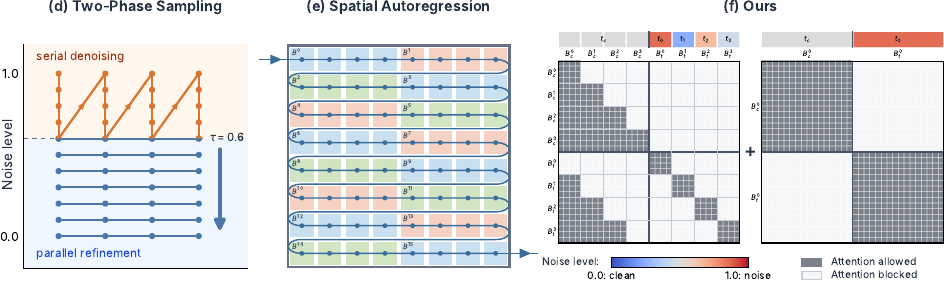}
  \caption{\textbf{Attention patterns across methods and our sampling strategy.}
  (a--c) Attention masks and noise levels for Bidirectional, cDF, and BC.
  (d) Serial denoising followed by parallel refinement.
  (e) Token order within a temporal slice.
  (f) Multiple block sizes used by Ours during training.}
  \label{fig:method}
\end{figure}

\paragraph{Attention masks and noise levels.}
As shown in Figure~\ref{fig:method}a, bidirectional diffusion uses standard self-attention and one noise level
for the entire video. Each video token can attend to all video tokens during
training and the entire sequence is denoised in parallel during inference.
As shown in Figure~\ref{fig:method}b, Diffusion
Forcing~\citep{chen2024diffusionforcing} uses independent noise levels for
each block during training. This enables autoregressive generation
because the model learns to denoise each block using the surrounding partially
noised context. We implement a causal version of Diffusion Forcing
Transformer~\citep{song2025history}, which we refer to as cDF, to better suit
autoregressive generation. We replace the default self-attention with
block-causal attention. Block-causal attention restricts
each block to attend only to itself and preceding blocks.
As shown in Figure~\ref{fig:method}c, block-causal diffusion~\citep{arriola2025blockdiffusion}, which we refer to as BC,
uses two copies of
the token sequence: a context copy
with one noise level \(t_c\) and a target copy with independent noise levels \(t_k\).
We denote context and target blocks, \(B_c^k\) and \(B_t^k\), respectively.
Each context block attends to itself and preceding context blocks. Each target
block attends to itself and preceding context blocks, but not other target blocks.
Loss is computed only on the target blocks.

\subsection{Sampling}

\method{} generates blocks autoregressively at high noise before refining
the whole video together in parallel at low noise. As shown in Figure~\ref{fig:method}d, we define a transition noise level
\(\tau\in[0,1]\) to separate the two phases.
By default, we use ten uniformly spaced denoising steps and \(\tau=0.6\) to allocate
four denoising steps to serial denoising and six to parallel refinement.
Let \(N_f\) denote the number of tokens in one temporal slice of the video token
grid and \(N_v\) the number of tokens in the full video. We use a block size
\(b\leq N_f\) for serial denoising and \(b=N_v\) for parallel refinement.

\paragraph{Serial denoising.}
We initialize blocks of size \(b\) with Gaussian noise and denoise them autoregressively from
\(t=1\) to \(t=\tau\). Each block is conditioned on a
key--value (KV) cache of previously generated blocks at noise level \(\tau\).
We follow the token order shown in Figure~\ref{fig:method}e.
For cDF and BC, we always use a block size of \(N_f\) and fully denoise each
block to \(t=0\). We re-noise the block before adding it to the KV cache to
mitigate autoregressive drift, using context noise level \(t_c=0.2\)
as default.

\paragraph{Parallel refinement.}
For \method{}, once all blocks reach \(\tau\), we treat the full noisy video as a single
block and denoise it from \(t=\tau\) to \(t=0\).
Despite using the same number of denoising steps per token, parallel refinement is much
faster than serial denoising because all tokens are batched together.
This amortizes the cost of reading model weights from memory. Sampling
is usually limited by memory bandwidth, not compute.

\subsection{Training}

\method{} learns to denoise each target block using only the preceding context
blocks at noise level \(t_c\) as context. Setting \(t_c=\tau\) would match the context
noise level used during training with the transition noise level used during sampling.
By default, we train with \(t_c\sim\mathcal{U}(0,1)\) to support different transition
noise levels at inference time. We sample the noise level for each target block
\(t_k\) from a plateau logit-normal
distribution~\citep{bfl2025representation}. This distribution preserves training coverage
at high noise levels.

As shown in Figure~\ref{fig:method}f, \method{} uses the same token sequence,
attention pattern, and noise levels as block-causal diffusion. The only difference
is that we train with multiple block sizes. We mix samples with different block sizes within the same
training batch. Training with the block size mixture \(b\in\{N_f,N_v\}\)
would support \method{} sampling at a single block size. We use
\(b\in\{4,8,\ldots,N_f,N_v\}\) to support flexible block sizes. We delay the
choice of both block size and transition noise level to inference time.
To support image-to-video generation, we prepend the clean condition-image
tokens to the token sequence. The image tokens are omitted from the attention
patterns in Figure~\ref{fig:method} for clarity.

We implement all four methods---Bidirectional, cDF, BC, and \method{}---with two
architectures: DiT-B and Wan-5B-LoRA. DiT-B is a 132.7M-parameter diffusion
transformer~\citep{peebles2023dit} trained from scratch in pixel space. Wan-5B-LoRA is a pretrained
latent-space video diffusion model adapted using rank-32 LoRA~\citep{hu2022lora}
with 80.6M trainable parameters in total. In Wan-5B, we replace
the default attention mask with each method's custom attention pattern and let
the adapters adapt.
Since BC and \method{} use double the sequence length during training, we
double the batch size used by Bidirectional and cDF to compensate. This matches training token budgets
and roughly matches total training FLOPs. For more implementation details,
see Appendix~\ref{app:implementation}.

\section{Datasets}
\label{sec:dataset}

\inlinepanel{figure}{0.42\textwidth}{\begin{figure}[tbp]
  \centering
  \includegraphics[width=\linewidth]{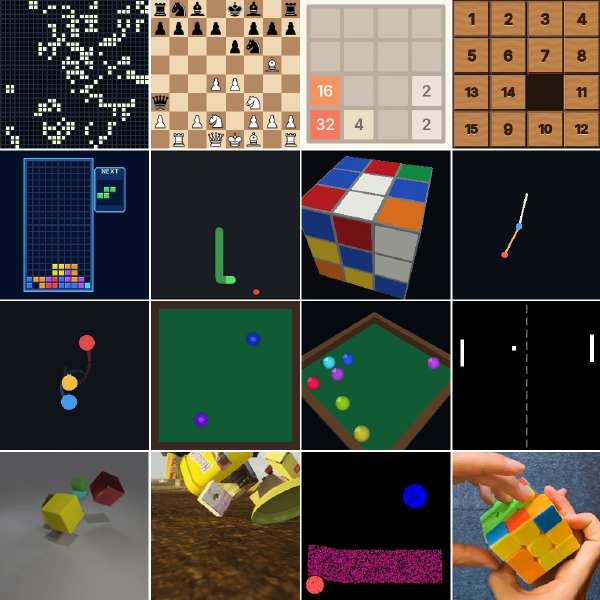}
  \caption{\textbf{Samples from all datasets.} Games, simulations, and real video.}
  \label{fig:mosaic}
\end{figure}
}[16]
Our dataset collection spans games, simulations and real video.
For simple datasets, we implement video parsing and symbolic evaluation
to detect rule and physics violations. For more complex datasets, we rely
on distributional metrics based on Fr\'echet distance~\citep{dowson1982frechet}.

\subsection{Dataset Collection}

First, we implement Conway's Game of Life~\citep{gardner1970life}, a
zero-player game where a grid of live and dead cells evolves deterministically.
At each step, all cells update simultaneously according to their neighbors.
A live cell survives if it has two or three live neighbors and dies otherwise.
A dead cell comes to life if it has exactly three live neighbors.
To create our video dataset, we hold each step for \(p_t\) frames,
aligning state transitions with token boundaries.

Next, we add other discrete datasets like Chess, 2048,
15 Puzzle, Tetris, Snake,
and Rubik's Cube 3D.
We animate transitions between states to make the dynamics easier
for a video model to learn. Animations occupy the
first \(p_t-1\) frames of each temporal block but are not used during
evaluation. We parse only the settled frame at the end of the state transition.
If the game ends early, we overlay an end-of-game modal and retain it for the
rest of the video.

\WFclear
\inlinepanel[r]{table}{0.42\textwidth}{%
  \renewenvironment{tabular}[1]
    {\begin{tabular*}{\linewidth}{@{\hskip\tabcolsep\extracolsep{\fill}}#1}}
    {\end{tabular*}}
  \begin{table}[tbp]
  \centering
  \begin{threeparttable}
  \caption{\textbf{Dataset pool sizes.} All videos are 10 seconds at 24 fps and \(256\times256\).}
  \label{tab:data-spec}
  \begin{tabular}{lcc}
    \toprule
    Dataset & Train pool & Val pool \\
    \midrule
    \rowcolor{black!6} \multicolumn{3}{l}{\textit{Discrete datasets}} \\
    All & \(\sim\!\infty\) & \(\sim\!\infty\) \\
    \rowcolor{black!6} \multicolumn{3}{l}{\textit{Continuous datasets}} \\
    All & \(\sim\!\infty\) & \(\sim\!\infty\) \\
    \rowcolor{black!6} \multicolumn{3}{l}{\textit{Video datasets}} \\
    Kubric MOVi-A & 9,703 & 250 \\
    Kubric MOVi-C & 9,737 & 250 \\
    MPMWorlds & 9,439 & 1,017 \\
    Rubik's Cube Real & 5,000 & 512 \\
    \bottomrule
  \end{tabular}%
  \end{threeparttable}
\end{table}

}
Then, we add continuous simulations to test physical dynamics.
We implement Double Pendulum, three-body orbital dynamics, 2D and 3D colliding balls,
and Pong.
Since we evaluate for valid transitions rather than optimal play for games,
we either sample random legal moves or use bots to produce reasonable move
sequences. For Chess, we downloaded 1M human-played chess
games from Lichess. All datasets so far are rendered on-the-fly during training.

Lastly, we include datasets with more complex simulations and real video.
These datasets are pre-generated offline, downloaded with their official splits,
or collected ourselves.
We use Kubric MOVi-A and MOVi-C~\citep{greff2022kubric} to simulate 3D colliding
objects at two different levels of complexity.
We add MPMWorlds~\citep{kovacic2026mpmworlds}, which uses 2D material point
method simulation with deformable objects, fluids, and moving obstacles to test
dynamics involving deformation and fluid flow.
We collect Rubik's Cube Real, a dataset containing 30 minutes of real footage of
scrambling and solving a Rubik's Cube. We divide the footage into non-overlapping
training and validation segments, then extract heavily overlapping 10-second
clips within each split.

\subsection{Video Parsing and Evaluation}

\inlinepanel{figure}{0.42\textwidth}{%
  \begin{figure}[tbp]
  \centering
  \includegraphics[width=\linewidth]{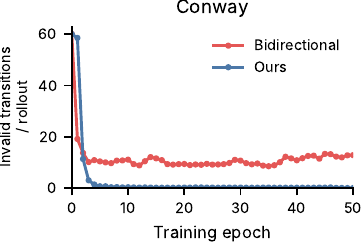}
  \caption{\textbf{Learning Conway rules.} Ours rapidly learns valid transitions, while the baseline continues to make errors.}
  \label{fig:training-dynamics}
\end{figure}

  \par\medskip
  \begin{figure}[tbp]
  \centering
  \includegraphics[width=\linewidth]{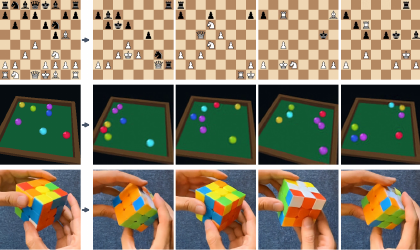}
  \caption{\textbf{One image, multiple continuations.} Ours produces multiple plausible continuations from the same image.}
  \label{fig:diversity}
\end{figure}

}[27]

For simple datasets, we evaluate generated videos by parsing them into symbolic states and checking whether
their transitions obey the underlying game rules or physical dynamics.
Discrete datasets evolve through a finite set of transitions on a fixed grid.
Since the grid positions are known ahead of time, we parse their videos
through template matching of their grid cells. Then, we evaluate each pair independently, checking state validity and the
permitted changes between frames. Persistent invalid states incur repeated penalties. We count \emph{invalid transitions per rollout}
as our primary metric.
For Rubik's Cube 3D, since the camera pose is unknown, we fit a 3D cube to
the 2D silhouette before parsing the facelet colors.
For continuous datasets, we track expected objects and fit simulation parameters. Then, we account for occlusions and measure deviations from the expected dynamics. We report cumulative error as \emph{dynamics error per rollout}.
We use red diagnostic overlays to localize the errors we detect and verify each
dataset evaluation procedure manually on generated samples. We invite the reader
to use our project page to do the same. Parsing and tracking failures complicate
our evaluation metrics. For more details, see Appendix~\ref{app:datasets}.

With these metrics, Figure~\ref{fig:training-dynamics} shows that the bidirectional
baseline never learns to generate valid transitions for Conway over
the course of training. In contrast, our method
rapidly learns the rules and generates videos with almost no errors.
Our video parsing and metrics do not require a pixel-aligned
ground-truth video for evaluation. We evaluate each sample against the underlying
rules and dynamics of the simulator. Figure~\ref{fig:diversity} shows that our
method produces diverse valid continuations from a single image. However, for video
datasets that are too complex to parse reliably, we fall back to using
distributional metrics like FVD. We adopt CD-FVD~\citep{ge2024contentbias} as our primary
distributional metric because it uses a more modern feature extractor and
has been shown to have less bias toward per-frame appearance and greater
sensitivity to temporal distortions than FVD.
Nonetheless, we also report FVD in Appendix~\ref{app:fvd}.

\begin{table}[tbp]
  \caption{\textbf{Main comparisons.} Serial methods outperform the bidirectional baseline and perform similarly to each other across all datasets. cDF and BC use block sizes of 64. Ours uses a block size of 8 for discrete datasets and 64 for continuous and video datasets. Best is marked in blue and worst is marked in red.}
  \label{tab:comparison}
  \centering
  \resizebox{\linewidth}{!}{%
    \dimen0=0pt%
    \settowidth{\dimen2}{9.4}\ifdim\dimen2>\dimen0 \dimen0=\dimen2\fi%
    \settowidth{\dimen2}{51.0}\ifdim\dimen2>\dimen0 \dimen0=\dimen2\fi%
    \settowidth{\dimen2}{47.9}\ifdim\dimen2>\dimen0 \dimen0=\dimen2\fi%
    \settowidth{\dimen2}{59.1}\ifdim\dimen2>\dimen0 \dimen0=\dimen2\fi%
    \settowidth{\dimen2}{57.2}\ifdim\dimen2>\dimen0 \dimen0=\dimen2\fi%
    \settowidth{\dimen2}{47.2}\ifdim\dimen2>\dimen0 \dimen0=\dimen2\fi%
    \settowidth{\dimen2}{46.1}\ifdim\dimen2>\dimen0 \dimen0=\dimen2\fi%
    \settowidth{\dimen2}{73.4}\ifdim\dimen2>\dimen0 \dimen0=\dimen2\fi%
    \settowidth{\dimen2}{59.5}\ifdim\dimen2>\dimen0 \dimen0=\dimen2\fi%
    \settowidth{\dimen2}{5.3}\ifdim\dimen2>\dimen0 \dimen0=\dimen2\fi%
    \settowidth{\dimen2}{6.3}\ifdim\dimen2>\dimen0 \dimen0=\dimen2\fi%
    \settowidth{\dimen2}{159.7}\ifdim\dimen2>\dimen0 \dimen0=\dimen2\fi%
    \settowidth{\dimen2}{252.8}\ifdim\dimen2>\dimen0 \dimen0=\dimen2\fi%
    \settowidth{\dimen2}{512.4}\ifdim\dimen2>\dimen0 \dimen0=\dimen2\fi%
    \settowidth{\dimen2}{88.7}\ifdim\dimen2>\dimen0 \dimen0=\dimen2\fi%
    \settowidth{\dimen2}{698.7}\ifdim\dimen2>\dimen0 \dimen0=\dimen2\fi%
    \expandafter\edef\csname metricwidth3\endcsname{\the\dimen0}%
    \dimen0=0pt%
    \settowidth{\dimen2}{0.5}\ifdim\dimen2>\dimen0 \dimen0=\dimen2\fi%
    \settowidth{\dimen2}{4.6}\ifdim\dimen2>\dimen0 \dimen0=\dimen2\fi%
    \settowidth{\dimen2}{1.5}\ifdim\dimen2>\dimen0 \dimen0=\dimen2\fi%
    \settowidth{\dimen2}{21.0}\ifdim\dimen2>\dimen0 \dimen0=\dimen2\fi%
    \settowidth{\dimen2}{4.5}\ifdim\dimen2>\dimen0 \dimen0=\dimen2\fi%
    \settowidth{\dimen2}{1.4}\ifdim\dimen2>\dimen0 \dimen0=\dimen2\fi%
    \settowidth{\dimen2}{26.5}\ifdim\dimen2>\dimen0 \dimen0=\dimen2\fi%
    \settowidth{\dimen2}{1.2}\ifdim\dimen2>\dimen0 \dimen0=\dimen2\fi%
    \settowidth{\dimen2}{0.4}\ifdim\dimen2>\dimen0 \dimen0=\dimen2\fi%
    \settowidth{\dimen2}{0.0}\ifdim\dimen2>\dimen0 \dimen0=\dimen2\fi%
    \settowidth{\dimen2}{0.1}\ifdim\dimen2>\dimen0 \dimen0=\dimen2\fi%
    \settowidth{\dimen2}{168.5}\ifdim\dimen2>\dimen0 \dimen0=\dimen2\fi%
    \settowidth{\dimen2}{376.5}\ifdim\dimen2>\dimen0 \dimen0=\dimen2\fi%
    \settowidth{\dimen2}{43.2}\ifdim\dimen2>\dimen0 \dimen0=\dimen2\fi%
    \settowidth{\dimen2}{70.8}\ifdim\dimen2>\dimen0 \dimen0=\dimen2\fi%
    \expandafter\edef\csname metricwidth4\endcsname{\the\dimen0}%
    \dimen0=0pt%
    \settowidth{\dimen2}{0.0}\ifdim\dimen2>\dimen0 \dimen0=\dimen2\fi%
    \settowidth{\dimen2}{6.8}\ifdim\dimen2>\dimen0 \dimen0=\dimen2\fi%
    \settowidth{\dimen2}{1.4}\ifdim\dimen2>\dimen0 \dimen0=\dimen2\fi%
    \settowidth{\dimen2}{7.3}\ifdim\dimen2>\dimen0 \dimen0=\dimen2\fi%
    \settowidth{\dimen2}{5.3}\ifdim\dimen2>\dimen0 \dimen0=\dimen2\fi%
    \settowidth{\dimen2}{31.1}\ifdim\dimen2>\dimen0 \dimen0=\dimen2\fi%
    \settowidth{\dimen2}{1.2}\ifdim\dimen2>\dimen0 \dimen0=\dimen2\fi%
    \settowidth{\dimen2}{0.2}\ifdim\dimen2>\dimen0 \dimen0=\dimen2\fi%
    \settowidth{\dimen2}{0.1}\ifdim\dimen2>\dimen0 \dimen0=\dimen2\fi%
    \settowidth{\dimen2}{141.2}\ifdim\dimen2>\dimen0 \dimen0=\dimen2\fi%
    \settowidth{\dimen2}{240.3}\ifdim\dimen2>\dimen0 \dimen0=\dimen2\fi%
    \settowidth{\dimen2}{43.9}\ifdim\dimen2>\dimen0 \dimen0=\dimen2\fi%
    \settowidth{\dimen2}{70.9}\ifdim\dimen2>\dimen0 \dimen0=\dimen2\fi%
    \expandafter\edef\csname metricwidth5\endcsname{\the\dimen0}%
    \dimen0=0pt%
    \settowidth{\dimen2}{0.0}\ifdim\dimen2>\dimen0 \dimen0=\dimen2\fi%
    \settowidth{\dimen2}{12.3}\ifdim\dimen2>\dimen0 \dimen0=\dimen2\fi%
    \settowidth{\dimen2}{0.9}\ifdim\dimen2>\dimen0 \dimen0=\dimen2\fi%
    \settowidth{\dimen2}{12.7}\ifdim\dimen2>\dimen0 \dimen0=\dimen2\fi%
    \settowidth{\dimen2}{3.2}\ifdim\dimen2>\dimen0 \dimen0=\dimen2\fi%
    \settowidth{\dimen2}{2.4}\ifdim\dimen2>\dimen0 \dimen0=\dimen2\fi%
    \settowidth{\dimen2}{22.0}\ifdim\dimen2>\dimen0 \dimen0=\dimen2\fi%
    \settowidth{\dimen2}{1.1}\ifdim\dimen2>\dimen0 \dimen0=\dimen2\fi%
    \settowidth{\dimen2}{0.6}\ifdim\dimen2>\dimen0 \dimen0=\dimen2\fi%
    \settowidth{\dimen2}{1.3}\ifdim\dimen2>\dimen0 \dimen0=\dimen2\fi%
    \settowidth{\dimen2}{137.0}\ifdim\dimen2>\dimen0 \dimen0=\dimen2\fi%
    \settowidth{\dimen2}{231.7}\ifdim\dimen2>\dimen0 \dimen0=\dimen2\fi%
    \settowidth{\dimen2}{42.5}\ifdim\dimen2>\dimen0 \dimen0=\dimen2\fi%
    \settowidth{\dimen2}{64.4}\ifdim\dimen2>\dimen0 \dimen0=\dimen2\fi%
    \expandafter\edef\csname metricwidth6\endcsname{\the\dimen0}%
    \begin{tabular}{lcccccc}
      \toprule
      \multicolumn{3}{c}{} & Parallel & \multicolumn{3}{c}{Serial} \\
      \cmidrule(lr){4-4} \cmidrule(lr){5-7}
      Dataset & Architecture & Metric & Bidirectional & cDF & BC & Ours \\
      \midrule
      \rowcolor{black!6} \multicolumn{7}{l}{\textit{Discrete datasets}} \\
      Conway & \multirow{7}{*}{DiT-B} & \multirow{7}{*}{\(\dfrac{\text{Invalid transitions}}{\text{rollout}}\;\text{\large\(\downarrow\)}\)} & {\setlength{\fboxsep}{0pt}\colorbox{red!18}{\hspace{1.5pt}\makebox[\csname metricwidth3\endcsname][c]{\strut 9.4}\hspace{1.5pt}}} & 0.5 & {\setlength{\fboxsep}{0pt}\colorbox{blue!18}{\hspace{1.5pt}\makebox[\csname metricwidth5\endcsname][c]{\strut 0.0}\hspace{1.5pt}}} & {\setlength{\fboxsep}{0pt}\colorbox{blue!18}{\hspace{1.5pt}\makebox[\csname metricwidth6\endcsname][c]{\strut 0.0}\hspace{1.5pt}}} \\
      Chess &  &  & {\setlength{\fboxsep}{0pt}\colorbox{red!18}{\hspace{1.5pt}\makebox[\csname metricwidth3\endcsname][c]{\strut 51.0}\hspace{1.5pt}}} & {\setlength{\fboxsep}{0pt}\colorbox{blue!18}{\hspace{1.5pt}\makebox[\csname metricwidth4\endcsname][c]{\strut 4.6}\hspace{1.5pt}}} & 6.8 & 12.3 \\
      2048 &  &  & {\setlength{\fboxsep}{0pt}\colorbox{red!18}{\hspace{1.5pt}\makebox[\csname metricwidth3\endcsname][c]{\strut 47.9}\hspace{1.5pt}}} & 1.5 & 1.4 & {\setlength{\fboxsep}{0pt}\colorbox{blue!18}{\hspace{1.5pt}\makebox[\csname metricwidth6\endcsname][c]{\strut 0.9}\hspace{1.5pt}}} \\
      Fifteen Puzzle &  &  & {\setlength{\fboxsep}{0pt}\colorbox{red!18}{\hspace{1.5pt}\makebox[\csname metricwidth3\endcsname][c]{\strut 59.1}\hspace{1.5pt}}} & 21.0 & {\setlength{\fboxsep}{0pt}\colorbox{blue!18}{\hspace{1.5pt}\makebox[\csname metricwidth5\endcsname][c]{\strut 7.3}\hspace{1.5pt}}} & 12.7 \\
      Tetris &  &  & {\setlength{\fboxsep}{0pt}\colorbox{red!18}{\hspace{1.5pt}\makebox[\csname metricwidth3\endcsname][c]{\strut 57.2}\hspace{1.5pt}}} & 4.5 & 5.3 & {\setlength{\fboxsep}{0pt}\colorbox{blue!18}{\hspace{1.5pt}\makebox[\csname metricwidth6\endcsname][c]{\strut 3.2}\hspace{1.5pt}}} \\
      Snake &  &  & {\setlength{\fboxsep}{0pt}\colorbox{red!18}{\hspace{1.5pt}\makebox[\csname metricwidth3\endcsname][c]{\strut 47.2}\hspace{1.5pt}}} & {\setlength{\fboxsep}{0pt}\colorbox{blue!18}{\hspace{1.5pt}\makebox[\csname metricwidth4\endcsname][c]{\strut 1.4}\hspace{1.5pt}}} & {\setlength{\fboxsep}{0pt}\colorbox{blue!18}{\hspace{1.5pt}\makebox[\csname metricwidth5\endcsname][c]{\strut 1.4}\hspace{1.5pt}}} & 2.4 \\
      Rubik's Cube 3D &  &  & {\setlength{\fboxsep}{0pt}\colorbox{red!18}{\hspace{1.5pt}\makebox[\csname metricwidth3\endcsname][c]{\strut 46.1}\hspace{1.5pt}}} & 26.5 & 31.1 & {\setlength{\fboxsep}{0pt}\colorbox{blue!18}{\hspace{1.5pt}\makebox[\csname metricwidth6\endcsname][c]{\strut 22.0}\hspace{1.5pt}}} \\
      \midrule
      \rowcolor{black!6} \multicolumn{7}{l}{\textit{Continuous datasets}} \\
      Double Pendulum & \multirow{5}{*}{DiT-B} & \multirow{5}{*}{\(\dfrac{\text{Dynamics error}}{\text{rollout}}\;\text{\large\(\downarrow\)}\)} & {\setlength{\fboxsep}{0pt}\colorbox{red!18}{\hspace{1.5pt}\makebox[\csname metricwidth3\endcsname][c]{\strut 73.4}\hspace{1.5pt}}} & 1.2 & 1.2 & {\setlength{\fboxsep}{0pt}\colorbox{blue!18}{\hspace{1.5pt}\makebox[\csname metricwidth6\endcsname][c]{\strut 1.1}\hspace{1.5pt}}} \\
      Three Body &  &  & {\setlength{\fboxsep}{0pt}\colorbox{red!18}{\hspace{1.5pt}\makebox[\csname metricwidth3\endcsname][c]{\strut 59.5}\hspace{1.5pt}}} & 0.4 & {\setlength{\fboxsep}{0pt}\colorbox{blue!18}{\hspace{1.5pt}\makebox[\csname metricwidth5\endcsname][c]{\strut 0.2}\hspace{1.5pt}}} & 0.6 \\
      Colliding Balls 2D &  &  & {\setlength{\fboxsep}{0pt}\colorbox{red!18}{\hspace{1.5pt}\makebox[\csname metricwidth3\endcsname][c]{\strut 5.3}\hspace{1.5pt}}} & {\setlength{\fboxsep}{0pt}\colorbox{blue!18}{\hspace{1.5pt}\makebox[\csname metricwidth4\endcsname][c]{\strut 0.0}\hspace{1.5pt}}} & {\setlength{\fboxsep}{0pt}\colorbox{blue!18}{\hspace{1.5pt}\makebox[\csname metricwidth5\endcsname][c]{\strut 0.0}\hspace{1.5pt}}} & {\setlength{\fboxsep}{0pt}\colorbox{blue!18}{\hspace{1.5pt}\makebox[\csname metricwidth6\endcsname][c]{\strut 0.0}\hspace{1.5pt}}} \\
      Colliding Balls 3D &  &  & {\setlength{\fboxsep}{0pt}\colorbox{red!18}{\hspace{1.5pt}\makebox[\csname metricwidth3\endcsname][c]{\strut 6.3}\hspace{1.5pt}}} & 0.1 & 0.1 & {\setlength{\fboxsep}{0pt}\colorbox{blue!18}{\hspace{1.5pt}\makebox[\csname metricwidth6\endcsname][c]{\strut 0.0}\hspace{1.5pt}}} \\
      Pong &  &  & {\setlength{\fboxsep}{0pt}\colorbox{red!18}{\hspace{1.5pt}\makebox[\csname metricwidth3\endcsname][c]{\strut 159.7}\hspace{1.5pt}}} & {\setlength{\fboxsep}{0pt}\colorbox{blue!18}{\hspace{1.5pt}\makebox[\csname metricwidth4\endcsname][c]{\strut 0.1}\hspace{1.5pt}}} & 0.2 & 1.3 \\
      \midrule
      \rowcolor{black!6} \multicolumn{7}{l}{\textit{Video datasets}} \\
      Kubric MOVi-A & \multirow{4}{*}{Wan-5B-LoRA} & \multirow{4}{*}{CD-FVD \(\downarrow\)} & {\setlength{\fboxsep}{0pt}\colorbox{red!18}{\hspace{1.5pt}\makebox[\csname metricwidth3\endcsname][c]{\strut 252.8}\hspace{1.5pt}}} & 168.5 & 141.2 & {\setlength{\fboxsep}{0pt}\colorbox{blue!18}{\hspace{1.5pt}\makebox[\csname metricwidth6\endcsname][c]{\strut 137.0}\hspace{1.5pt}}} \\
      Kubric MOVi-C &  &  & {\setlength{\fboxsep}{0pt}\colorbox{red!18}{\hspace{1.5pt}\makebox[\csname metricwidth3\endcsname][c]{\strut 512.4}\hspace{1.5pt}}} & 376.5 & 240.3 & {\setlength{\fboxsep}{0pt}\colorbox{blue!18}{\hspace{1.5pt}\makebox[\csname metricwidth6\endcsname][c]{\strut 231.7}\hspace{1.5pt}}} \\
      MPMWorlds &  &  & {\setlength{\fboxsep}{0pt}\colorbox{red!18}{\hspace{1.5pt}\makebox[\csname metricwidth3\endcsname][c]{\strut 88.7}\hspace{1.5pt}}} & 43.2 & 43.9 & {\setlength{\fboxsep}{0pt}\colorbox{blue!18}{\hspace{1.5pt}\makebox[\csname metricwidth6\endcsname][c]{\strut 42.5}\hspace{1.5pt}}} \\
      Rubik's Cube Real &  &  & {\setlength{\fboxsep}{0pt}\colorbox{red!18}{\hspace{1.5pt}\makebox[\csname metricwidth3\endcsname][c]{\strut 698.7}\hspace{1.5pt}}} & 70.8 & 70.9 & {\setlength{\fboxsep}{0pt}\colorbox{blue!18}{\hspace{1.5pt}\makebox[\csname metricwidth6\endcsname][c]{\strut 64.4}\hspace{1.5pt}}} \\
      \bottomrule
    \end{tabular}%
  }
\end{table}

\begin{figure}[t]
  \centering
  \includegraphics[width=\linewidth]{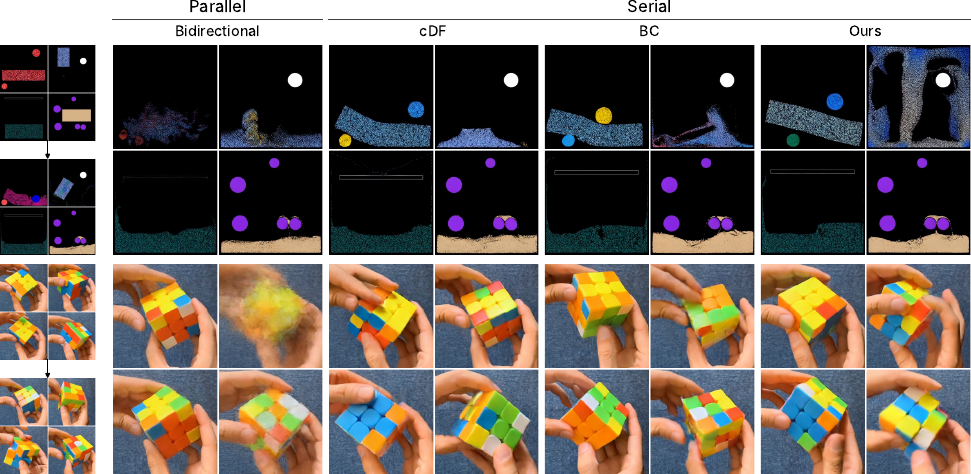}
  \caption{\textbf{Qualitative comparisons with the Wan-5B-LoRA architecture.} Serial methods perform similarly and produce more coherent video than the bidirectional baseline on MPMWorlds and Rubik's Cube Real datasets.}
  \label{fig:results-examples}
\end{figure}

\section{Results}
\label{sec:results}

We train a dedicated model for each dataset and method. As shown in
Table~\ref{tab:comparison}, all serial methods perform similarly on discrete and
continuous datasets with no clear winner. However, they dramatically
outperform the bidirectional baseline.
Figure~\ref{fig:teaser} shows still images from video samples generated by our method
and the baseline. Diagnostic overlays highlight errors in red.
The baseline frequently makes invalid chess moves and 2048 transitions.
It loses balls in simulations and produces unrealistic Pong gameplay.
While not perfect, our method, and all serial methods in fact, generate much
more coherent video than the baseline, with the same model size. We show
uncurated samples for all methods in the video gallery on our project page.
Serial generation continues to outperform the parallel baseline on more
complex datasets with a larger architecture. Figure~\ref{fig:results-examples}
shows that the standard configuration of LoRA on Wan-5B generates implausible material
deformations on MPMWorlds. Matched serial counterparts generate much more
reasonable continuations and perform much better on Rubik's Cube Real.

\inlinepanel[r]{table}{0pt}{%
  \fontsize{8.5}{10}\selectfont
  \begin{table}[tbp]
  \centering
  \begin{threeparttable}
  \caption{\textbf{CD-FVD \(\downarrow\) with Wan-5B-LoRA at block size 64.} Sampling times are shown in parentheses.}
  \label{tab:video-block-size}
  \dimen0=0pt%
  \settowidth{\dimen2}{168.5}\ifdim\dimen2>\dimen0 \dimen0=\dimen2\fi%
  \settowidth{\dimen2}{376.5}\ifdim\dimen2>\dimen0 \dimen0=\dimen2\fi%
  \settowidth{\dimen2}{43.2}\ifdim\dimen2>\dimen0 \dimen0=\dimen2\fi%
  \settowidth{\dimen2}{70.8}\ifdim\dimen2>\dimen0 \dimen0=\dimen2\fi%
  \expandafter\edef\csname metricwidth2\endcsname{\the\dimen0}%
  \dimen0=0pt%
  \settowidth{\dimen2}{141.2}\ifdim\dimen2>\dimen0 \dimen0=\dimen2\fi%
  \settowidth{\dimen2}{240.3}\ifdim\dimen2>\dimen0 \dimen0=\dimen2\fi%
  \settowidth{\dimen2}{43.9}\ifdim\dimen2>\dimen0 \dimen0=\dimen2\fi%
  \settowidth{\dimen2}{70.9}\ifdim\dimen2>\dimen0 \dimen0=\dimen2\fi%
  \expandafter\edef\csname metricwidth3\endcsname{\the\dimen0}%
  \dimen0=0pt%
  \settowidth{\dimen2}{137.0}\ifdim\dimen2>\dimen0 \dimen0=\dimen2\fi%
  \settowidth{\dimen2}{231.7}\ifdim\dimen2>\dimen0 \dimen0=\dimen2\fi%
  \settowidth{\dimen2}{42.5}\ifdim\dimen2>\dimen0 \dimen0=\dimen2\fi%
  \settowidth{\dimen2}{64.4}\ifdim\dimen2>\dimen0 \dimen0=\dimen2\fi%
  \expandafter\edef\csname metricwidth4\endcsname{\the\dimen0}%
  \begin{tabular}{l|ccc}
    \toprule
     & cDF & BC & Ours \\
    \cmidrule(lr){2-2} \cmidrule(lr){3-3} \cmidrule(lr){4-4}
    \multicolumn{1}{r|}{Block size} & 64 & 64 & 64 \\
    \midrule
    \rowcolor{black!6} \multicolumn{4}{l}{\textit{Video datasets}} \\
    Kubric MOVi-A & 168.5 (13.8\ s) & 141.2 (13.7\ s) & {\setlength{\fboxsep}{0pt}\colorbox{blue!18}{\hspace{1.5pt}\makebox[\csname metricwidth4\endcsname][c]{\strut 137.0}\hspace{1.5pt}}} (6.8\ s) \\
    Kubric MOVi-C & 376.5 (13.7\ s) & 240.3 (13.8\ s) & {\setlength{\fboxsep}{0pt}\colorbox{blue!18}{\hspace{1.5pt}\makebox[\csname metricwidth4\endcsname][c]{\strut 231.7}\hspace{1.5pt}}} (6.8\ s) \\
    MPMWorlds & 43.2 (13.7\ s) & 43.9 (13.6\ s) & {\setlength{\fboxsep}{0pt}\colorbox{blue!18}{\hspace{1.5pt}\makebox[\csname metricwidth4\endcsname][c]{\strut 42.5}\hspace{1.5pt}}} (6.9\ s) \\
    Rubik's Cube Real & 70.8 (13.8\ s) & 70.9 (13.7\ s) & {\setlength{\fboxsep}{0pt}\colorbox{blue!18}{\hspace{1.5pt}\makebox[\csname metricwidth4\endcsname][c]{\strut 64.4}\hspace{1.5pt}}} (7.0\ s) \\
    \bottomrule
  \end{tabular}%
  \end{threeparttable}
\end{table}

}[11]
\paragraph{Sampling efficiency.}
With the same block size, Table~\ref{tab:video-block-size} shows that \method{}
achieves the lowest CD-FVD at twice the speed compared to other serial methods.
Table~\ref{tab:transition-noise} shows that for \method{}, a transition noise level
\(\tau=0.6\) is more than twice as fast as \(\tau=0.0\) across all block sizes.

\begin{figure}[t]
  \centering
  \includegraphics[width=\linewidth]{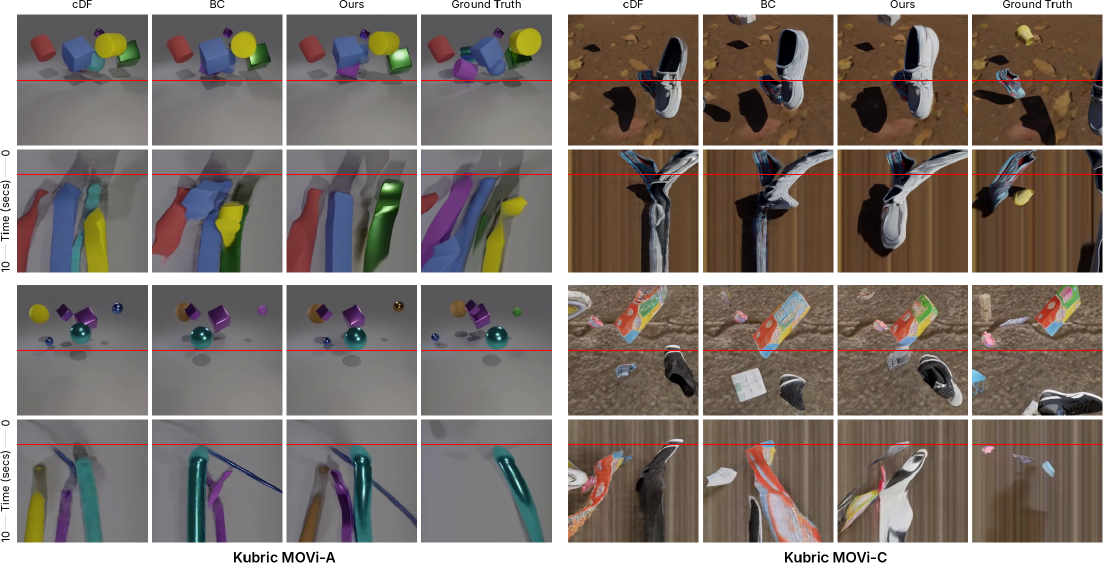}
  \caption{\textbf{Spatiotemporal slices expose temporal flickering.} Ours and ground truth show clean vertical lines while cDF and BC show subtle patchy artifacts.}
  \label{fig:flicker}
\end{figure}

\Needspace{6\baselineskip}
\paragraph{Temporal stability.}
\label{sec:main-comparison}
For Kubric datasets, purely serial generation shows
temporal instability or flicker.
This is best viewed on the project website.
Figure~\ref{fig:flicker} shows \(x\)--\(t\) spatiotemporal slices formed by
stacking the horizontal center line of each frame vertically. Our method and
ground truth show clean vertical lines, particularly
in the static background, whereas BC and cDF show subtle patchy artifacts.
Our parallel refinement stage lets all blocks reconcile their appearance,
smoothing artifacts and achieving the lowest CD-FVD on all four video datasets.

\subsection{Ablations}
\label{sec:ablations}

We vary the transition noise level, block size, and token order to test how the amount and
organization of serial computation affect generation quality and sampling time.
Unless specified otherwise, all ablations use DiT-B models trained from scratch
with block sizes \(b\in\{4,8,\ldots,N_f,N_v\}\) and sampled with a block size
\(b=8\) and a transition noise level \(\tau=0.6\).

\paragraph{Transition noise level.}
The transition noise level \(\tau\) determines how much denoising is performed
serially before parallel refinement begins. We can vary it at inference time
from fully parallel generation at \(\tau=1.0\) to fully serial generation at
\(\tau=0.0\). However,
\(\tau=0.0\) is not recommended because, at this setting, the model is highly
susceptible to autoregressive drift. Table~\ref{tab:transition-noise} shows
that \(\tau=0.6\) achieves a similar score to \(\tau=0.2\) while being
significantly faster. We use \(\tau=0.6\) as the default for all our datasets.
See Appendix~\ref{app:context-noise} for the full sweep of context noise levels
for all serial methods.

\paragraph{Block size.}
Table~\ref{tab:block-size-mixture}
shows that smaller block sizes reduce errors for discrete datasets.
Continuous datasets seem less sensitive to the choice of block size.
We train additional sets of models called Pin-64 with block sizes
\(b \in \{64, N_v\}\) and Pin-8 with block sizes \(b \in \{8, N_v\}\).
Focusing training on a particular block size improves performance for that block
size at the cost of flexibility.

\paragraph{Token order.}
\label{sec:token-order}

For small block sizes, the token order determines the spatial grouping used by
individual blocks. We compare Raster, Zigzag, and Hilbert
orderings in Table~\ref{tab:token-order}. Zigzag uses an order similar to
JPEG~\citep{wallace1991jpeg}, and Hilbert uses a 2D Hilbert
curve~\citep{hilbert1891curve} starting in the top-left corner.
More elaborate token orders offer no advantage over Raster, so we use it as default.

\begin{table}[t]
  \begingroup
  \renewenvironment{table}[1][]{}{}
  \begin{table}[!t]
  \caption{\textbf{Balancing serial denoising and parallel refinement.} Moderate transition noise levels perform similarly. Small blocks improve performance but increase sampling cost. We report \emph{invalid transitions per rollout} \(\downarrow\) with sampling time in parentheses. Default settings are marked in grey.}
  \label{tab:transition-noise}
  \centering
  \resizebox{\linewidth}{!}{%
    \begin{tabular}{r|cccc|cccc}
      \toprule
       & \multicolumn{4}{c}{Chess} & \multicolumn{4}{c}{2048} \\
      \cmidrule(lr){2-5} \cmidrule(lr){6-9}
      Block size & 64 & 32 & 16 & \cellcolor{black!18}8 & 64 & 32 & 16 & \cellcolor{black!18}8 \\
      \midrule
      \textit{fully parallel}\quad 1.0 & 52.1 (0.1\ s) & 52.1 (0.1\ s) & 52.1 (0.1\ s) & \cellcolor{black!18}52.1 (0.1\ s) & 48.5 (0.1\ s) & 48.5 (0.1\ s) & 48.5 (0.1\ s) & \cellcolor{black!18}48.5 (0.1\ s) \\
      0.8 & 31.7 (1.0\ s) & 28.3 (1.9\ s) & 17.1 (3.6\ s) & \cellcolor{black!18}15.3 (6.4\ s) & 11.4 (0.9\ s) & 8.3 (1.7\ s) & 4.7 (3.3\ s) & \cellcolor{black!18}3.1 (5.4\ s) \\
      \cellcolor{black!18}\(\tau=0.6\) & \cellcolor{black!18}24.6 (1.6\ s) & \cellcolor{black!18}24.6 (3.0\ s) & \cellcolor{black!18}15.1 (5.9\ s) & \cellcolor{black!30}13.3 (10.4\ s) & \cellcolor{black!18}4.5 (1.5\ s) & \cellcolor{black!18}3.0 (2.9\ s) & \cellcolor{black!18}1.3 (5.7\ s) & \cellcolor{black!30}1.0 (9.3\ s) \\
      0.4 & 24.0 (2.2\ s) & 24.8 (4.2\ s) & 15.1 (8.3\ s) & \cellcolor{black!18}13.9 (13.4\ s) & 4.4 (2.1\ s) & 2.7 (4.0\ s) & 1.5 (7.9\ s) & \cellcolor{black!18}1.0 (12.7\ s) \\
      0.2 & 24.6 (2.8\ s) & 23.5 (5.4\ s) & 14.1 (10.7\ s) & \cellcolor{black!18}13.1 (17.2\ s) & 4.3 (2.7\ s) & 2.7 (5.1\ s) & 1.6 (10.1\ s) & \cellcolor{black!18}1.1 (16.5\ s) \\
      \textit{fully serial}\quad 0.0 & 34.0 (3.4\ s) & 32.5 (6.5\ s) & 27.1 (13.0\ s) & \cellcolor{black!18}26.3 (21.0\ s) & 6.3 (3.1\ s) & 5.5 (6.1\ s) & 3.3 (12.0\ s) & \cellcolor{black!18}2.7 (19.2\ s) \\
      \bottomrule
    \end{tabular}%
  }
\end{table}

  \par\medskip
  \fontsize{8.5}{10}\selectfont
  \setlength{\tabcolsep}{2pt}
  \renewenvironment{tabular}[1]
    {\begin{tabular*}{\linewidth}{@{\hskip\tabcolsep\extracolsep{\fill}}#1}}
    {\end{tabular*}}
  \noindent
  \begin{minipage}[t]{0.56\linewidth}
    \vspace{0pt}
    \begin{table}[tbp]
  \centering
  \begin{threeparttable}
  \caption{\textbf{Block size ablation.} Decreasing block size reduces errors more for discrete datasets than continuous datasets. Training for a specific block size mostly improves discrete datasets.}
  \label{tab:block-size-mixture}
  \dimen0=0pt%
  \settowidth{\dimen2}{0.0}\ifdim\dimen2>\dimen0 \dimen0=\dimen2\fi%
  \settowidth{\dimen2}{8.9}\ifdim\dimen2>\dimen0 \dimen0=\dimen2\fi%
  \settowidth{\dimen2}{1.8}\ifdim\dimen2>\dimen0 \dimen0=\dimen2\fi%
  \settowidth{\dimen2}{26.4}\ifdim\dimen2>\dimen0 \dimen0=\dimen2\fi%
  \settowidth{\dimen2}{7.0}\ifdim\dimen2>\dimen0 \dimen0=\dimen2\fi%
  \settowidth{\dimen2}{1.5}\ifdim\dimen2>\dimen0 \dimen0=\dimen2\fi%
  \settowidth{\dimen2}{36.5}\ifdim\dimen2>\dimen0 \dimen0=\dimen2\fi%
  \settowidth{\dimen2}{1.1}\ifdim\dimen2>\dimen0 \dimen0=\dimen2\fi%
  \settowidth{\dimen2}{0.1}\ifdim\dimen2>\dimen0 \dimen0=\dimen2\fi%
  \settowidth{\dimen2}{0.2}\ifdim\dimen2>\dimen0 \dimen0=\dimen2\fi%
  \expandafter\edef\csname metricwidth3\endcsname{\the\dimen0}%
  \dimen0=0pt%
  \settowidth{\dimen2}{0.0}\ifdim\dimen2>\dimen0 \dimen0=\dimen2\fi%
  \settowidth{\dimen2}{25.2}\ifdim\dimen2>\dimen0 \dimen0=\dimen2\fi%
  \settowidth{\dimen2}{4.4}\ifdim\dimen2>\dimen0 \dimen0=\dimen2\fi%
  \settowidth{\dimen2}{54.9}\ifdim\dimen2>\dimen0 \dimen0=\dimen2\fi%
  \settowidth{\dimen2}{10.0}\ifdim\dimen2>\dimen0 \dimen0=\dimen2\fi%
  \settowidth{\dimen2}{6.3}\ifdim\dimen2>\dimen0 \dimen0=\dimen2\fi%
  \settowidth{\dimen2}{36.9}\ifdim\dimen2>\dimen0 \dimen0=\dimen2\fi%
  \settowidth{\dimen2}{1.1}\ifdim\dimen2>\dimen0 \dimen0=\dimen2\fi%
  \settowidth{\dimen2}{0.6}\ifdim\dimen2>\dimen0 \dimen0=\dimen2\fi%
  \settowidth{\dimen2}{1.3}\ifdim\dimen2>\dimen0 \dimen0=\dimen2\fi%
  \expandafter\edef\csname metricwidth4\endcsname{\the\dimen0}%
  \dimen0=0pt%
  \settowidth{\dimen2}{0.0}\ifdim\dimen2>\dimen0 \dimen0=\dimen2\fi%
  \settowidth{\dimen2}{24.4}\ifdim\dimen2>\dimen0 \dimen0=\dimen2\fi%
  \settowidth{\dimen2}{2.9}\ifdim\dimen2>\dimen0 \dimen0=\dimen2\fi%
  \settowidth{\dimen2}{48.5}\ifdim\dimen2>\dimen0 \dimen0=\dimen2\fi%
  \settowidth{\dimen2}{9.6}\ifdim\dimen2>\dimen0 \dimen0=\dimen2\fi%
  \settowidth{\dimen2}{4.7}\ifdim\dimen2>\dimen0 \dimen0=\dimen2\fi%
  \settowidth{\dimen2}{31.0}\ifdim\dimen2>\dimen0 \dimen0=\dimen2\fi%
  \settowidth{\dimen2}{1.1}\ifdim\dimen2>\dimen0 \dimen0=\dimen2\fi%
  \settowidth{\dimen2}{0.3}\ifdim\dimen2>\dimen0 \dimen0=\dimen2\fi%
  \settowidth{\dimen2}{0.7}\ifdim\dimen2>\dimen0 \dimen0=\dimen2\fi%
  \expandafter\edef\csname metricwidth5\endcsname{\the\dimen0}%
  \dimen0=0pt%
  \settowidth{\dimen2}{0.0}\ifdim\dimen2>\dimen0 \dimen0=\dimen2\fi%
  \settowidth{\dimen2}{15.4}\ifdim\dimen2>\dimen0 \dimen0=\dimen2\fi%
  \settowidth{\dimen2}{1.4}\ifdim\dimen2>\dimen0 \dimen0=\dimen2\fi%
  \settowidth{\dimen2}{16.8}\ifdim\dimen2>\dimen0 \dimen0=\dimen2\fi%
  \settowidth{\dimen2}{2.8}\ifdim\dimen2>\dimen0 \dimen0=\dimen2\fi%
  \settowidth{\dimen2}{3.6}\ifdim\dimen2>\dimen0 \dimen0=\dimen2\fi%
  \settowidth{\dimen2}{26.9}\ifdim\dimen2>\dimen0 \dimen0=\dimen2\fi%
  \settowidth{\dimen2}{1.1}\ifdim\dimen2>\dimen0 \dimen0=\dimen2\fi%
  \settowidth{\dimen2}{0.2}\ifdim\dimen2>\dimen0 \dimen0=\dimen2\fi%
  \settowidth{\dimen2}{0.1}\ifdim\dimen2>\dimen0 \dimen0=\dimen2\fi%
  \settowidth{\dimen2}{0.4}\ifdim\dimen2>\dimen0 \dimen0=\dimen2\fi%
  \expandafter\edef\csname metricwidth6\endcsname{\the\dimen0}%
  \dimen0=0pt%
  \settowidth{\dimen2}{0.0}\ifdim\dimen2>\dimen0 \dimen0=\dimen2\fi%
  \settowidth{\dimen2}{12.3}\ifdim\dimen2>\dimen0 \dimen0=\dimen2\fi%
  \settowidth{\dimen2}{0.9}\ifdim\dimen2>\dimen0 \dimen0=\dimen2\fi%
  \settowidth{\dimen2}{12.7}\ifdim\dimen2>\dimen0 \dimen0=\dimen2\fi%
  \settowidth{\dimen2}{3.2}\ifdim\dimen2>\dimen0 \dimen0=\dimen2\fi%
  \settowidth{\dimen2}{2.4}\ifdim\dimen2>\dimen0 \dimen0=\dimen2\fi%
  \settowidth{\dimen2}{22.0}\ifdim\dimen2>\dimen0 \dimen0=\dimen2\fi%
  \settowidth{\dimen2}{1.1}\ifdim\dimen2>\dimen0 \dimen0=\dimen2\fi%
  \settowidth{\dimen2}{0.3}\ifdim\dimen2>\dimen0 \dimen0=\dimen2\fi%
  \settowidth{\dimen2}{0.2}\ifdim\dimen2>\dimen0 \dimen0=\dimen2\fi%
  \expandafter\edef\csname metricwidth7\endcsname{\the\dimen0}%
  \dimen0=0pt%
  \settowidth{\dimen2}{0.0}\ifdim\dimen2>\dimen0 \dimen0=\dimen2\fi%
  \settowidth{\dimen2}{10.3}\ifdim\dimen2>\dimen0 \dimen0=\dimen2\fi%
  \settowidth{\dimen2}{0.8}\ifdim\dimen2>\dimen0 \dimen0=\dimen2\fi%
  \settowidth{\dimen2}{3.4}\ifdim\dimen2>\dimen0 \dimen0=\dimen2\fi%
  \settowidth{\dimen2}{4.5}\ifdim\dimen2>\dimen0 \dimen0=\dimen2\fi%
  \settowidth{\dimen2}{2.6}\ifdim\dimen2>\dimen0 \dimen0=\dimen2\fi%
  \settowidth{\dimen2}{21.6}\ifdim\dimen2>\dimen0 \dimen0=\dimen2\fi%
  \settowidth{\dimen2}{1.1}\ifdim\dimen2>\dimen0 \dimen0=\dimen2\fi%
  \settowidth{\dimen2}{0.5}\ifdim\dimen2>\dimen0 \dimen0=\dimen2\fi%
  \settowidth{\dimen2}{0.1}\ifdim\dimen2>\dimen0 \dimen0=\dimen2\fi%
  \expandafter\edef\csname metricwidth8\endcsname{\the\dimen0}%
  \begin{tabular}{l|cccccc}
    \toprule
     & Pin & \multicolumn{4}{c}{Flex} & Pin \\
    \cmidrule(lr){2-2} \cmidrule(lr){3-6} \cmidrule(lr){7-7}
    \multicolumn{1}{r|}{Block size} & 64 & 64 & 32 & 16 & 8 & 8 \\
    \midrule
    \rowcolor{black!6} \multicolumn{7}{l}{\textit{Discrete datasets}} \\
    \makebox[85pt][l]{Conway} & {\setlength{\fboxsep}{0pt}\colorbox{blue!18}{\hspace{1.5pt}\makebox[\csname metricwidth3\endcsname][c]{\strut 0.0}\hspace{1.5pt}}} & {\setlength{\fboxsep}{0pt}\colorbox{blue!18}{\hspace{1.5pt}\makebox[\csname metricwidth4\endcsname][c]{\strut 0.0}\hspace{1.5pt}}} & {\setlength{\fboxsep}{0pt}\colorbox{blue!18}{\hspace{1.5pt}\makebox[\csname metricwidth5\endcsname][c]{\strut 0.0}\hspace{1.5pt}}} & {\setlength{\fboxsep}{0pt}\colorbox{blue!18}{\hspace{1.5pt}\makebox[\csname metricwidth6\endcsname][c]{\strut 0.0}\hspace{1.5pt}}} & {\setlength{\fboxsep}{0pt}\colorbox{blue!18}{\hspace{1.5pt}\makebox[\csname metricwidth7\endcsname][c]{\strut 0.0}\hspace{1.5pt}}} & {\setlength{\fboxsep}{0pt}\colorbox{blue!18}{\hspace{1.5pt}\makebox[\csname metricwidth8\endcsname][c]{\strut 0.0}\hspace{1.5pt}}} \\
    \makebox[85pt][l]{Chess} & {\setlength{\fboxsep}{0pt}\colorbox{blue!18}{\hspace{1.5pt}\makebox[\csname metricwidth3\endcsname][c]{\strut 8.9}\hspace{1.5pt}}} & 25.2 & 24.4 & 15.4 & 12.3 & {\setlength{\fboxsep}{0pt}\colorbox{blue!8}{\hspace{1.5pt}\makebox[\csname metricwidth8\endcsname][c]{\strut 10.3}\hspace{1.5pt}}} \\
    \makebox[85pt][l]{2048} & 1.8 & 4.4 & 2.9 & 1.4 & {\setlength{\fboxsep}{0pt}\colorbox{blue!8}{\hspace{1.5pt}\makebox[\csname metricwidth7\endcsname][c]{\strut 0.9}\hspace{1.5pt}}} & {\setlength{\fboxsep}{0pt}\colorbox{blue!18}{\hspace{1.5pt}\makebox[\csname metricwidth8\endcsname][c]{\strut 0.8}\hspace{1.5pt}}} \\
    \makebox[85pt][l]{Fifteen Puzzle} & 26.4 & 54.9 & 48.5 & 16.8 & {\setlength{\fboxsep}{0pt}\colorbox{blue!8}{\hspace{1.5pt}\makebox[\csname metricwidth7\endcsname][c]{\strut 12.7}\hspace{1.5pt}}} & {\setlength{\fboxsep}{0pt}\colorbox{blue!18}{\hspace{1.5pt}\makebox[\csname metricwidth8\endcsname][c]{\strut 3.4}\hspace{1.5pt}}} \\
    \makebox[85pt][l]{Tetris} & 7.0 & 10.0 & 9.6 & {\setlength{\fboxsep}{0pt}\colorbox{blue!18}{\hspace{1.5pt}\makebox[\csname metricwidth6\endcsname][c]{\strut 2.8}\hspace{1.5pt}}} & {\setlength{\fboxsep}{0pt}\colorbox{blue!8}{\hspace{1.5pt}\makebox[\csname metricwidth7\endcsname][c]{\strut 3.2}\hspace{1.5pt}}} & 4.5 \\
    \makebox[85pt][l]{Snake} & {\setlength{\fboxsep}{0pt}\colorbox{blue!18}{\hspace{1.5pt}\makebox[\csname metricwidth3\endcsname][c]{\strut 1.5}\hspace{1.5pt}}} & 6.3 & 4.7 & 3.6 & {\setlength{\fboxsep}{0pt}\colorbox{blue!8}{\hspace{1.5pt}\makebox[\csname metricwidth7\endcsname][c]{\strut 2.4}\hspace{1.5pt}}} & 2.6 \\
    \makebox[85pt][l]{Rubik's Cube 3D} & 36.5 & 36.9 & 31.0 & 26.9 & {\setlength{\fboxsep}{0pt}\colorbox{blue!8}{\hspace{1.5pt}\makebox[\csname metricwidth7\endcsname][c]{\strut 22.0}\hspace{1.5pt}}} & {\setlength{\fboxsep}{0pt}\colorbox{blue!18}{\hspace{1.5pt}\makebox[\csname metricwidth8\endcsname][c]{\strut 21.6}\hspace{1.5pt}}} \\
    \midrule
    \rowcolor{black!6} \multicolumn{7}{l}{\textit{Continuous datasets}} \\
    \makebox[85pt][l]{Double Pendulum} & {\setlength{\fboxsep}{0pt}\colorbox{blue!18}{\hspace{1.5pt}\makebox[\csname metricwidth3\endcsname][c]{\strut 1.1}\hspace{1.5pt}}} & {\setlength{\fboxsep}{0pt}\colorbox{blue!18}{\hspace{1.5pt}\makebox[\csname metricwidth4\endcsname][c]{\strut 1.1}\hspace{1.5pt}}} & {\setlength{\fboxsep}{0pt}\colorbox{blue!18}{\hspace{1.5pt}\makebox[\csname metricwidth5\endcsname][c]{\strut 1.1}\hspace{1.5pt}}} & {\setlength{\fboxsep}{0pt}\colorbox{blue!18}{\hspace{1.5pt}\makebox[\csname metricwidth6\endcsname][c]{\strut 1.1}\hspace{1.5pt}}} & {\setlength{\fboxsep}{0pt}\colorbox{blue!18}{\hspace{1.5pt}\makebox[\csname metricwidth7\endcsname][c]{\strut 1.1}\hspace{1.5pt}}} & {\setlength{\fboxsep}{0pt}\colorbox{blue!18}{\hspace{1.5pt}\makebox[\csname metricwidth8\endcsname][c]{\strut 1.1}\hspace{1.5pt}}} \\
    \makebox[85pt][l]{Three Body} & {\setlength{\fboxsep}{0pt}\colorbox{blue!18}{\hspace{1.5pt}\makebox[\csname metricwidth3\endcsname][c]{\strut 0.1}\hspace{1.5pt}}} & 0.6 & 0.3 & {\setlength{\fboxsep}{0pt}\colorbox{blue!8}{\hspace{1.5pt}\makebox[\csname metricwidth6\endcsname][c]{\strut 0.2}\hspace{1.5pt}}} & 0.3 & 0.5 \\
    \makebox[85pt][l]{Colliding Balls 2D} & {\setlength{\fboxsep}{0pt}\colorbox{blue!18}{\hspace{1.5pt}\makebox[\csname metricwidth3\endcsname][c]{\strut 0.0}\hspace{1.5pt}}} & {\setlength{\fboxsep}{0pt}\colorbox{blue!18}{\hspace{1.5pt}\makebox[\csname metricwidth4\endcsname][c]{\strut 0.0}\hspace{1.5pt}}} & {\setlength{\fboxsep}{0pt}\colorbox{blue!18}{\hspace{1.5pt}\makebox[\csname metricwidth5\endcsname][c]{\strut 0.0}\hspace{1.5pt}}} & {\setlength{\fboxsep}{0pt}\colorbox{blue!18}{\hspace{1.5pt}\makebox[\csname metricwidth6\endcsname][c]{\strut 0.0}\hspace{1.5pt}}} & {\setlength{\fboxsep}{0pt}\colorbox{blue!18}{\hspace{1.5pt}\makebox[\csname metricwidth7\endcsname][c]{\strut 0.0}\hspace{1.5pt}}} & {\setlength{\fboxsep}{0pt}\colorbox{blue!18}{\hspace{1.5pt}\makebox[\csname metricwidth8\endcsname][c]{\strut 0.0}\hspace{1.5pt}}} \\
    \makebox[85pt][l]{Colliding Balls 3D} & 0.1 & {\setlength{\fboxsep}{0pt}\colorbox{blue!18}{\hspace{1.5pt}\makebox[\csname metricwidth4\endcsname][c]{\strut 0.0}\hspace{1.5pt}}} & {\setlength{\fboxsep}{0pt}\colorbox{blue!18}{\hspace{1.5pt}\makebox[\csname metricwidth5\endcsname][c]{\strut 0.0}\hspace{1.5pt}}} & 0.1 & {\setlength{\fboxsep}{0pt}\colorbox{blue!18}{\hspace{1.5pt}\makebox[\csname metricwidth7\endcsname][c]{\strut 0.0}\hspace{1.5pt}}} & 0.1 \\
    \makebox[85pt][l]{Pong} & {\setlength{\fboxsep}{0pt}\colorbox{blue!8}{\hspace{1.5pt}\makebox[\csname metricwidth3\endcsname][c]{\strut 0.2}\hspace{1.5pt}}} & 1.3 & 0.7 & 0.4 & {\setlength{\fboxsep}{0pt}\colorbox{blue!8}{\hspace{1.5pt}\makebox[\csname metricwidth7\endcsname][c]{\strut 0.2}\hspace{1.5pt}}} & {\setlength{\fboxsep}{0pt}\colorbox{blue!18}{\hspace{1.5pt}\makebox[\csname metricwidth8\endcsname][c]{\strut 0.1}\hspace{1.5pt}}} \\
    \bottomrule
  \end{tabular}%
  \end{threeparttable}
\end{table}

  \end{minipage}\hspace{0.01\linewidth}%
  \begin{minipage}[t]{0.43\linewidth}
    \vspace{0pt}
    \begin{table}[tbp]
  \centering
  \begin{threeparttable}
  \caption{\textbf{Token order ablation.} Zigzag and Hilbert orders offer no consistent advantage over raster order, which we use as the default.}
  \label{tab:token-order}
  \dimen0=0pt%
  \settowidth{\dimen2}{0.0}\ifdim\dimen2>\dimen0 \dimen0=\dimen2\fi%
  \settowidth{\dimen2}{10.3}\ifdim\dimen2>\dimen0 \dimen0=\dimen2\fi%
  \settowidth{\dimen2}{0.8}\ifdim\dimen2>\dimen0 \dimen0=\dimen2\fi%
  \settowidth{\dimen2}{3.4}\ifdim\dimen2>\dimen0 \dimen0=\dimen2\fi%
  \settowidth{\dimen2}{4.5}\ifdim\dimen2>\dimen0 \dimen0=\dimen2\fi%
  \settowidth{\dimen2}{2.6}\ifdim\dimen2>\dimen0 \dimen0=\dimen2\fi%
  \settowidth{\dimen2}{21.6}\ifdim\dimen2>\dimen0 \dimen0=\dimen2\fi%
  \settowidth{\dimen2}{1.1}\ifdim\dimen2>\dimen0 \dimen0=\dimen2\fi%
  \settowidth{\dimen2}{0.5}\ifdim\dimen2>\dimen0 \dimen0=\dimen2\fi%
  \settowidth{\dimen2}{0.1}\ifdim\dimen2>\dimen0 \dimen0=\dimen2\fi%
  \expandafter\edef\csname metricwidth3\endcsname{\the\dimen0}%
  \dimen0=0pt%
  \settowidth{\dimen2}{0.0}\ifdim\dimen2>\dimen0 \dimen0=\dimen2\fi%
  \settowidth{\dimen2}{14.9}\ifdim\dimen2>\dimen0 \dimen0=\dimen2\fi%
  \settowidth{\dimen2}{1.3}\ifdim\dimen2>\dimen0 \dimen0=\dimen2\fi%
  \settowidth{\dimen2}{2.4}\ifdim\dimen2>\dimen0 \dimen0=\dimen2\fi%
  \settowidth{\dimen2}{6.3}\ifdim\dimen2>\dimen0 \dimen0=\dimen2\fi%
  \settowidth{\dimen2}{21.2}\ifdim\dimen2>\dimen0 \dimen0=\dimen2\fi%
  \settowidth{\dimen2}{1.2}\ifdim\dimen2>\dimen0 \dimen0=\dimen2\fi%
  \settowidth{\dimen2}{0.4}\ifdim\dimen2>\dimen0 \dimen0=\dimen2\fi%
  \settowidth{\dimen2}{0.1}\ifdim\dimen2>\dimen0 \dimen0=\dimen2\fi%
  \settowidth{\dimen2}{0.2}\ifdim\dimen2>\dimen0 \dimen0=\dimen2\fi%
  \expandafter\edef\csname metricwidth4\endcsname{\the\dimen0}%
  \dimen0=0pt%
  \settowidth{\dimen2}{0.0}\ifdim\dimen2>\dimen0 \dimen0=\dimen2\fi%
  \settowidth{\dimen2}{11.8}\ifdim\dimen2>\dimen0 \dimen0=\dimen2\fi%
  \settowidth{\dimen2}{1.0}\ifdim\dimen2>\dimen0 \dimen0=\dimen2\fi%
  \settowidth{\dimen2}{1.6}\ifdim\dimen2>\dimen0 \dimen0=\dimen2\fi%
  \settowidth{\dimen2}{10.3}\ifdim\dimen2>\dimen0 \dimen0=\dimen2\fi%
  \settowidth{\dimen2}{4.1}\ifdim\dimen2>\dimen0 \dimen0=\dimen2\fi%
  \settowidth{\dimen2}{22.8}\ifdim\dimen2>\dimen0 \dimen0=\dimen2\fi%
  \settowidth{\dimen2}{1.2}\ifdim\dimen2>\dimen0 \dimen0=\dimen2\fi%
  \settowidth{\dimen2}{0.7}\ifdim\dimen2>\dimen0 \dimen0=\dimen2\fi%
  \settowidth{\dimen2}{0.1}\ifdim\dimen2>\dimen0 \dimen0=\dimen2\fi%
  \settowidth{\dimen2}{0.3}\ifdim\dimen2>\dimen0 \dimen0=\dimen2\fi%
  \expandafter\edef\csname metricwidth5\endcsname{\the\dimen0}%
  \begin{tabular}{l|ccc}
    \toprule
     & Raster & Zigzag & Hilbert \\
    \cmidrule(lr){2-2} \cmidrule(lr){3-3} \cmidrule(lr){4-4}
    \multicolumn{1}{r|}{Block size} & 8 & 8 & 8 \\
    \midrule
    \rowcolor{black!6} \multicolumn{4}{l}{\textit{Discrete datasets}} \\
    \makebox[85pt][l]{Conway} & {\setlength{\fboxsep}{0pt}\colorbox{blue!18}{\hspace{1.5pt}\makebox[\csname metricwidth3\endcsname][c]{\strut 0.0}\hspace{1.5pt}}} & {\setlength{\fboxsep}{0pt}\colorbox{blue!18}{\hspace{1.5pt}\makebox[\csname metricwidth4\endcsname][c]{\strut 0.0}\hspace{1.5pt}}} & {\setlength{\fboxsep}{0pt}\colorbox{blue!18}{\hspace{1.5pt}\makebox[\csname metricwidth5\endcsname][c]{\strut 0.0}\hspace{1.5pt}}} \\
    \makebox[85pt][l]{Chess} & {\setlength{\fboxsep}{0pt}\colorbox{blue!18}{\hspace{1.5pt}\makebox[\csname metricwidth3\endcsname][c]{\strut 10.3}\hspace{1.5pt}}} & 14.9 & {\setlength{\fboxsep}{0pt}\colorbox{blue!8}{\hspace{1.5pt}\makebox[\csname metricwidth5\endcsname][c]{\strut 11.8}\hspace{1.5pt}}} \\
    \makebox[85pt][l]{2048} & {\setlength{\fboxsep}{0pt}\colorbox{blue!18}{\hspace{1.5pt}\makebox[\csname metricwidth3\endcsname][c]{\strut 0.8}\hspace{1.5pt}}} & 1.3 & {\setlength{\fboxsep}{0pt}\colorbox{blue!8}{\hspace{1.5pt}\makebox[\csname metricwidth5\endcsname][c]{\strut 1.0}\hspace{1.5pt}}} \\
    \makebox[85pt][l]{Fifteen Puzzle} & 3.4 & {\setlength{\fboxsep}{0pt}\colorbox{blue!8}{\hspace{1.5pt}\makebox[\csname metricwidth4\endcsname][c]{\strut 2.4}\hspace{1.5pt}}} & {\setlength{\fboxsep}{0pt}\colorbox{blue!18}{\hspace{1.5pt}\makebox[\csname metricwidth5\endcsname][c]{\strut 1.6}\hspace{1.5pt}}} \\
    \makebox[85pt][l]{Tetris} & {\setlength{\fboxsep}{0pt}\colorbox{blue!18}{\hspace{1.5pt}\makebox[\csname metricwidth3\endcsname][c]{\strut 4.5}\hspace{1.5pt}}} & {\setlength{\fboxsep}{0pt}\colorbox{blue!8}{\hspace{1.5pt}\makebox[\csname metricwidth4\endcsname][c]{\strut 6.3}\hspace{1.5pt}}} & 10.3 \\
    \makebox[85pt][l]{Snake} & {\setlength{\fboxsep}{0pt}\colorbox{blue!8}{\hspace{1.5pt}\makebox[\csname metricwidth3\endcsname][c]{\strut 2.6}\hspace{1.5pt}}} & {\setlength{\fboxsep}{0pt}\colorbox{blue!18}{\hspace{1.5pt}\makebox[\csname metricwidth4\endcsname][c]{\strut 2.4}\hspace{1.5pt}}} & 4.1 \\
    \makebox[85pt][l]{Rubik's Cube 3D} & {\setlength{\fboxsep}{0pt}\colorbox{blue!8}{\hspace{1.5pt}\makebox[\csname metricwidth3\endcsname][c]{\strut 21.6}\hspace{1.5pt}}} & {\setlength{\fboxsep}{0pt}\colorbox{blue!18}{\hspace{1.5pt}\makebox[\csname metricwidth4\endcsname][c]{\strut 21.2}\hspace{1.5pt}}} & 22.8 \\
    \midrule
    \rowcolor{black!6} \multicolumn{4}{l}{\textit{Continuous datasets}} \\
    \makebox[85pt][l]{Double Pendulum} & {\setlength{\fboxsep}{0pt}\colorbox{blue!18}{\hspace{1.5pt}\makebox[\csname metricwidth3\endcsname][c]{\strut 1.1}\hspace{1.5pt}}} & {\setlength{\fboxsep}{0pt}\colorbox{blue!8}{\hspace{1.5pt}\makebox[\csname metricwidth4\endcsname][c]{\strut 1.2}\hspace{1.5pt}}} & {\setlength{\fboxsep}{0pt}\colorbox{blue!8}{\hspace{1.5pt}\makebox[\csname metricwidth5\endcsname][c]{\strut 1.2}\hspace{1.5pt}}} \\
    \makebox[85pt][l]{Three Body} & {\setlength{\fboxsep}{0pt}\colorbox{blue!8}{\hspace{1.5pt}\makebox[\csname metricwidth3\endcsname][c]{\strut 0.5}\hspace{1.5pt}}} & {\setlength{\fboxsep}{0pt}\colorbox{blue!18}{\hspace{1.5pt}\makebox[\csname metricwidth4\endcsname][c]{\strut 0.4}\hspace{1.5pt}}} & 0.7 \\
    \makebox[85pt][l]{Colliding Balls 2D} & {\setlength{\fboxsep}{0pt}\colorbox{blue!18}{\hspace{1.5pt}\makebox[\csname metricwidth3\endcsname][c]{\strut 0.0}\hspace{1.5pt}}} & {\setlength{\fboxsep}{0pt}\colorbox{blue!18}{\hspace{1.5pt}\makebox[\csname metricwidth4\endcsname][c]{\strut 0.0}\hspace{1.5pt}}} & {\setlength{\fboxsep}{0pt}\colorbox{blue!18}{\hspace{1.5pt}\makebox[\csname metricwidth5\endcsname][c]{\strut 0.0}\hspace{1.5pt}}} \\
    \makebox[85pt][l]{Colliding Balls 3D} & {\setlength{\fboxsep}{0pt}\colorbox{blue!18}{\hspace{1.5pt}\makebox[\csname metricwidth3\endcsname][c]{\strut 0.1}\hspace{1.5pt}}} & {\setlength{\fboxsep}{0pt}\colorbox{blue!18}{\hspace{1.5pt}\makebox[\csname metricwidth4\endcsname][c]{\strut 0.1}\hspace{1.5pt}}} & {\setlength{\fboxsep}{0pt}\colorbox{blue!18}{\hspace{1.5pt}\makebox[\csname metricwidth5\endcsname][c]{\strut 0.1}\hspace{1.5pt}}} \\
    \makebox[85pt][l]{Pong} & {\setlength{\fboxsep}{0pt}\colorbox{blue!18}{\hspace{1.5pt}\makebox[\csname metricwidth3\endcsname][c]{\strut 0.1}\hspace{1.5pt}}} & {\setlength{\fboxsep}{0pt}\colorbox{blue!8}{\hspace{1.5pt}\makebox[\csname metricwidth4\endcsname][c]{\strut 0.2}\hspace{1.5pt}}} & 0.3 \\
    \bottomrule
  \end{tabular}%
  \end{threeparttable}
\end{table}

  \end{minipage}
  \endgroup
\end{table}

\section{Conclusion}
\label{sec:conclusion}

The Serial Scaling Hypothesis~\citep{liu2026serialscaling} proved that diffusion
models are not very effective at modeling serial computation. Videos are
full of causal dependencies that require serial computation. Across a collection
of games and physical simulations, we demonstrate that scaling training
data does not enable bidirectional diffusion models to model complex video.
Across many datasets and two architectures, we show serial methods
outperform the parallel baseline with matched model sizes and training budgets.
We introduce \method{}, a simple extension to block-causal diffusion
that establishes video structure through autoregressive denoising at high noise
levels before refining the details in parallel. This preserves physical and logical
consistency from serial computation while improving sampling efficiency and temporal stability over
other serial methods.

\subsection*{AI use statement}

We used OpenAI Codex and Anthropic Claude for software implementation and
debugging, literature organization, experiment planning, and manuscript editing.
The authors reviewed the AI-assisted work and take responsibility for the
submission.

\ificlrfinal
\subsubsection*{Acknowledgments}

We thank Felix O'Mahony and Yuxin Yao for reviewing the manuscript and providing helpful feedback.
This work was funded in part by Toyota Motor Europe. The authors acknowledge
the use of resources provided by the Isambard-AI~\citep{mcintoshsmith2024isambard} National AI Research Resource
(AIRR). Isambard-AI is operated by the University of Bristol and is funded by
the UK Government's Department for Science, Innovation and Technology (DSIT)
via UK Research and Innovation; and the Science and Technology Facilities
Council [ST/AIRR/I-A-I/1023].
\fi

\clearpage
\appendix

\section{Implementation Details}
\label{app:implementation}
\label{app:evaluation}

\begin{table}[t]
  \caption{\textbf{Detailed hyperparameters for S2PD with DiT-B and Wan-LoRA-5B.} Shared entries span both columns. Architecture-specific parameters are listed in separate columns.}
  \label{tab:hyperparameters}
  \centering
  \resizebox{0.69\linewidth}{!}{%
    \begin{tabular}{lcc}
      \toprule
      Setting & DiT-B & Wan-5B-LoRA \\
      \midrule
      \rowcolor{black!6} \multicolumn{3}{l}{\textit{Architecture}} \\
      Depth, width, heads & 12, 768, 12 & 30, 3072, 24 \\
      Tubelet size & \(4\times32\times32\) pixels & \(1\times2\times2\) latents \\
      Objective & x-pred & v-pred \\
      Trainable parameters & 132.7M & 80.6M / 5.00B frozen \\
      Bottleneck dimension & 128 & --- \\
      LoRA rank / alpha & --- & 32 / 32 \\
      \midrule
      \rowcolor{black!6} \multicolumn{3}{l}{\textit{Training}} \\
      Loss & \multicolumn{2}{c}{v-loss} \\
      Epochs & \multicolumn{2}{c}{100} \\
      Steps per epoch & \multicolumn{2}{c}{1,000} \\
      Batch size & \multicolumn{2}{c}{32} \\
      LR schedule & \multicolumn{2}{c}{Cosine, \(10^{-4}\) to \(10^{-5}\)} \\
      Optimizer & \multicolumn{2}{c}{AdamW} \\
      Optimizer parameters & \(\beta_1=0.9,\ \beta_2=0.95\) & \(\beta_1=0.9,\ \beta_2=0.999\) \\
      Weight decay & 0 & 0.01 \\
      Gradient-norm clipping & \multicolumn{2}{c}{1.0} \\
      EMA decay & \multicolumn{2}{c}{0.9996} \\
      Context noise distribution & \multicolumn{2}{c}{\(\mathcal{U}(0,1)\)} \\
      Target noise distribution & \multicolumn{2}{c}{Plateau logit-normal, \(\mu=0.8,\ \sigma=0.8\)} \\
      Sequence length & \multicolumn{2}{c}{\(64+2\times3{,}840=7{,}744\) tokens} \\
      Block sizes & \multicolumn{2}{c}{4, 8, 16, 32, 64, 3,840} \\
      \midrule
      \rowcolor{black!6} \multicolumn{3}{l}{\textit{Sampling}} \\
      Transition noise level & \multicolumn{2}{c}{\(\tau=0.6\)} \\
      Solver & \multicolumn{2}{c}{Euler, 10 steps} \\
      Block size & 8 & 64 \\
      \bottomrule
    \end{tabular}%
  }
\end{table}

Table~\ref{tab:hyperparameters} summarizes architectural, training, and sampling
hyperparameters for DiT-B and Wan-5B-LoRA models.
For DiT-B, we use pixel-space diffusion with \(x\)-prediction and \(v\)-loss.
For Wan-5B-LoRA, we use latent-space diffusion with \(v\)-prediction and \(v\)-loss.
Both architectures encode text prompts with Wan's frozen UMT5-XXL encoder.
We project text token embeddings to the transformer's width and feed them into each transformer block via cross-attention.
We use a fixed prompt for each dataset except MPMWorlds. There, we use scene
descriptions provided by the dataset.
Training DiT-B takes about 12 hours on a single node with four H100 GPUs.
Training Wan-5B-LoRA takes about 48 hours on four nodes with four H100 GPUs each.
We train for 100 epochs with 1,000 steps per epoch.
To standardize training across finite and procedurally generated datasets,
we define one epoch as 1,000 optimizer steps and use batch sizes of 32 or 64,
depending on the method. For finite datasets, we repeat samples to fill epochs. We evaluate EMA checkpoints on validation splits. For datasets that are simulated on-the-fly, we use dedicated validation seeds
that are distinct from the training seeds.

\section{Dataset Details}
\label{app:datasets}
\label{app:scoring}

Let \(s_{t-1}\) and \(s_t\) be consecutive parsed states and
\(\mathcal{V}(s_{t-1})\) the permitted next states. We define the transition
residual as
\begin{equation}
  r_t=\min_{\hat{s}\in\mathcal{V}(s_{t-1})}d(\hat{s},s_t),
\end{equation}
where \(d\) measures how far the observed state \(s_t\) is from a permitted
state \(\hat{s}\).

\textbf{Discrete datasets.}
A transition is invalid when either state breaks the dataset's state rules or
no permitted move connects them. We report the number of invalid
transitions as \emph{invalid transitions per rollout}. A move is required between gameplay states.
Persistent state errors are charged on every transition they affect.

\textbf{Continuous datasets.} We extrapolate each frame from past frames using
the simulator's dynamics and compare that predicted extrapolation with what we
observe in the generated video.
For each object, let \(P_{t,i}\) and \(O_{t,i}\) be its predicted and observed
regions at time \(t\). Its residual is \(1-\operatorname{IoU}(P_{t,i},O_{t,i})\)
after a small positional tolerance for parsing error. We average the residual
over the source's objects and report the sum over frames as \emph{dynamics
error per rollout}. An expected object that is missing counts as a full residual error
for each frame that it is missing.

\subsection{Per-Dataset Details}

We describe the dataset-specific evaluation rules below. The released code provides the
full parsing procedures and details.

\textbf{Conway.} Each grid must follow from the previous grid under Conway's rules.

\textbf{Chess.} Each side must have one king, and pawns cannot stand on the first
or last rank. We infer the moving side from the first valid transition, then enforce
alternating turns. We infer possible castling
and en-passant rights from the visible board. Moving a pinned piece is illegal. A win modal
must follow a visible checkmate. We accept a draw modal after any valid board
without checkmate. An invalid modal counts
as one error when it appears.

\textbf{2048.} Each transition must consist of a horizontal or vertical slide
that changes the board, followed by the appearance of a new 2 or 4 tile in an
empty cell.

\textbf{Fifteen Puzzle.} Each board must contain every numbered tile and one
blank exactly once. A move slides a tile adjacent to the blank into that space.

\textbf{Tetris.} The previewed piece must drop vertically into a column and reach
its resting position when the board has no full rows. When the board contains
full rows, the next transition must clear them and shift the rows above downward,
leaving the preview unchanged. We allow any rotation of the dropped piece, and
the preview after a drop may show any piece.

\textbf{Snake.} A valid board contains one connected chain from head to tail
and one food cell. On each move, the head advances to a free neighboring cell
and the tail moves forward. If the head reaches food, the snake grows instead,
and new food appears in an empty cell. An end-of-game modal is valid only when
the preceding board is valid and the head is trapped.

\textbf{Rubik's Cube 3D.} We fit a cube to its silhouette and parse the colors
of the visible facelets. The silhouette determines the camera orientation only
up to a quarter turn. We therefore test the possible camera orientations and
legal quarter turns, accepting a transition if one combination matches every
facelet visible in both frames. We do not enforce color counts or solvability,
and hidden colors remain unconstrained. The metric thus checks local motion
rather than persistent identity through occlusion.

\textbf{Colliding Balls 2D.} We track balls by color and predict their motion
using the simulator's contact law, allowing collisions to occur between frames.
We penalize new overlaps between balls and new penetrations of the table rails.
Since balls cannot enter or leave the table, each unmatched arrival or departure
incurs one full ball penalty.

\textbf{Colliding Balls 3D.} We estimate the camera pose from the table and
apply the 2D checks in table coordinates. An arrival or departure is not
penalized when a nearby visible ball could explain it through occlusion, or when
the arriving or departing ball touches the image border. Each visible ball can
account for at most one arrival and one departure per frame. We exclude balls
clipped by the border from motion scoring and charge a failed table fit once
when it begins.

\textbf{Double Pendulum.} We fit rod lengths and the mass ratio once per
rollout, then use the pendulum equations to predict motion. Observed rod lengths
must remain close to their rollout medians. Unexplained missing bobs incur a
full-frame penalty. Only the second bob may leave the image without penalty.
Its last observations must place it near the edge or moving beyond it, and its
rod must be long enough to reach outside the image.

\textbf{Three Body.} We fit gravity once per rollout and predict
each body's motion from the forces exerted by visible bodies. When recent
motion indicates that a body has left the image or become occluded, we do not
penalize its absence until it returns. A body missing from the start is
penalized on every frame in which it is absent.

\textbf{Pong.} We predict ball motion using the simulator's wall and paddle
collisions. Paddles must remain on their rails and inside the court.
After crossing a goal line, the ball may reappear
at the serve position. Other missing or extra objects incur a penalty on every
affected frame.

\section{Additional Quantitative Results}
\label{app:supplementary-results}

\begin{samepage}
\begin{table}[H]
  \caption{\textbf{CD-FVD \(\downarrow\) with Wan-5B-LoRA at block size 64.} BC and cDF vary the re-noising used to resist autoregressive drift and Ours varies the transition noise level.}
  \label{tab:context-noise-cd}
  \centering
  \resizebox{\linewidth}{!}{%
    \dimen0=0pt%
    \settowidth{\dimen2}{176.1}\ifdim\dimen2>\dimen0 \dimen0=\dimen2\fi%
    \settowidth{\dimen2}{265.5}\ifdim\dimen2>\dimen0 \dimen0=\dimen2\fi%
    \settowidth{\dimen2}{48.2}\ifdim\dimen2>\dimen0 \dimen0=\dimen2\fi%
    \settowidth{\dimen2}{69.4}\ifdim\dimen2>\dimen0 \dimen0=\dimen2\fi%
    \expandafter\edef\csname metricwidth1\endcsname{\the\dimen0}%
    \dimen0=0pt%
    \settowidth{\dimen2}{141.2}\ifdim\dimen2>\dimen0 \dimen0=\dimen2\fi%
    \settowidth{\dimen2}{240.3}\ifdim\dimen2>\dimen0 \dimen0=\dimen2\fi%
    \settowidth{\dimen2}{43.9}\ifdim\dimen2>\dimen0 \dimen0=\dimen2\fi%
    \settowidth{\dimen2}{70.9}\ifdim\dimen2>\dimen0 \dimen0=\dimen2\fi%
    \expandafter\edef\csname metricwidth2\endcsname{\the\dimen0}%
    \dimen0=0pt%
    \settowidth{\dimen2}{153.8}\ifdim\dimen2>\dimen0 \dimen0=\dimen2\fi%
    \settowidth{\dimen2}{286.8}\ifdim\dimen2>\dimen0 \dimen0=\dimen2\fi%
    \settowidth{\dimen2}{49.0}\ifdim\dimen2>\dimen0 \dimen0=\dimen2\fi%
    \settowidth{\dimen2}{68.2}\ifdim\dimen2>\dimen0 \dimen0=\dimen2\fi%
    \expandafter\edef\csname metricwidth3\endcsname{\the\dimen0}%
    \dimen0=0pt%
    \settowidth{\dimen2}{182.6}\ifdim\dimen2>\dimen0 \dimen0=\dimen2\fi%
    \settowidth{\dimen2}{469.9}\ifdim\dimen2>\dimen0 \dimen0=\dimen2\fi%
    \settowidth{\dimen2}{63.3}\ifdim\dimen2>\dimen0 \dimen0=\dimen2\fi%
    \settowidth{\dimen2}{68.3}\ifdim\dimen2>\dimen0 \dimen0=\dimen2\fi%
    \expandafter\edef\csname metricwidth4\endcsname{\the\dimen0}%
    \dimen0=0pt%
    \settowidth{\dimen2}{278.7}\ifdim\dimen2>\dimen0 \dimen0=\dimen2\fi%
    \settowidth{\dimen2}{913.0}\ifdim\dimen2>\dimen0 \dimen0=\dimen2\fi%
    \settowidth{\dimen2}{115.5}\ifdim\dimen2>\dimen0 \dimen0=\dimen2\fi%
    \settowidth{\dimen2}{66.9}\ifdim\dimen2>\dimen0 \dimen0=\dimen2\fi%
    \expandafter\edef\csname metricwidth5\endcsname{\the\dimen0}%
    \dimen0=0pt%
    \settowidth{\dimen2}{366.2}\ifdim\dimen2>\dimen0 \dimen0=\dimen2\fi%
    \settowidth{\dimen2}{1353.9}\ifdim\dimen2>\dimen0 \dimen0=\dimen2\fi%
    \settowidth{\dimen2}{101.1}\ifdim\dimen2>\dimen0 \dimen0=\dimen2\fi%
    \settowidth{\dimen2}{96.2}\ifdim\dimen2>\dimen0 \dimen0=\dimen2\fi%
    \expandafter\edef\csname metricwidth6\endcsname{\the\dimen0}%
    \dimen0=0pt%
    \settowidth{\dimen2}{168.5}\ifdim\dimen2>\dimen0 \dimen0=\dimen2\fi%
    \settowidth{\dimen2}{376.5}\ifdim\dimen2>\dimen0 \dimen0=\dimen2\fi%
    \settowidth{\dimen2}{43.2}\ifdim\dimen2>\dimen0 \dimen0=\dimen2\fi%
    \settowidth{\dimen2}{70.8}\ifdim\dimen2>\dimen0 \dimen0=\dimen2\fi%
    \expandafter\edef\csname metricwidth7\endcsname{\the\dimen0}%
    \dimen0=0pt%
    \settowidth{\dimen2}{165.8}\ifdim\dimen2>\dimen0 \dimen0=\dimen2\fi%
    \settowidth{\dimen2}{345.4}\ifdim\dimen2>\dimen0 \dimen0=\dimen2\fi%
    \settowidth{\dimen2}{47.3}\ifdim\dimen2>\dimen0 \dimen0=\dimen2\fi%
    \settowidth{\dimen2}{71.9}\ifdim\dimen2>\dimen0 \dimen0=\dimen2\fi%
    \expandafter\edef\csname metricwidth8\endcsname{\the\dimen0}%
    \dimen0=0pt%
    \settowidth{\dimen2}{175.5}\ifdim\dimen2>\dimen0 \dimen0=\dimen2\fi%
    \settowidth{\dimen2}{454.1}\ifdim\dimen2>\dimen0 \dimen0=\dimen2\fi%
    \settowidth{\dimen2}{58.8}\ifdim\dimen2>\dimen0 \dimen0=\dimen2\fi%
    \settowidth{\dimen2}{71.8}\ifdim\dimen2>\dimen0 \dimen0=\dimen2\fi%
    \expandafter\edef\csname metricwidth9\endcsname{\the\dimen0}%
    \dimen0=0pt%
    \settowidth{\dimen2}{240.0}\ifdim\dimen2>\dimen0 \dimen0=\dimen2\fi%
    \settowidth{\dimen2}{894.9}\ifdim\dimen2>\dimen0 \dimen0=\dimen2\fi%
    \settowidth{\dimen2}{95.5}\ifdim\dimen2>\dimen0 \dimen0=\dimen2\fi%
    \settowidth{\dimen2}{74.9}\ifdim\dimen2>\dimen0 \dimen0=\dimen2\fi%
    \expandafter\edef\csname metricwidth10\endcsname{\the\dimen0}%
    \dimen0=0pt%
    \settowidth{\dimen2}{162.1}\ifdim\dimen2>\dimen0 \dimen0=\dimen2\fi%
    \settowidth{\dimen2}{319.9}\ifdim\dimen2>\dimen0 \dimen0=\dimen2\fi%
    \settowidth{\dimen2}{50.6}\ifdim\dimen2>\dimen0 \dimen0=\dimen2\fi%
    \settowidth{\dimen2}{68.6}\ifdim\dimen2>\dimen0 \dimen0=\dimen2\fi%
    \expandafter\edef\csname metricwidth11\endcsname{\the\dimen0}%
    \dimen0=0pt%
    \settowidth{\dimen2}{144.6}\ifdim\dimen2>\dimen0 \dimen0=\dimen2\fi%
    \settowidth{\dimen2}{250.4}\ifdim\dimen2>\dimen0 \dimen0=\dimen2\fi%
    \settowidth{\dimen2}{41.8}\ifdim\dimen2>\dimen0 \dimen0=\dimen2\fi%
    \settowidth{\dimen2}{66.6}\ifdim\dimen2>\dimen0 \dimen0=\dimen2\fi%
    \expandafter\edef\csname metricwidth12\endcsname{\the\dimen0}%
    \dimen0=0pt%
    \settowidth{\dimen2}{139.2}\ifdim\dimen2>\dimen0 \dimen0=\dimen2\fi%
    \settowidth{\dimen2}{227.0}\ifdim\dimen2>\dimen0 \dimen0=\dimen2\fi%
    \settowidth{\dimen2}{42.1}\ifdim\dimen2>\dimen0 \dimen0=\dimen2\fi%
    \settowidth{\dimen2}{65.6}\ifdim\dimen2>\dimen0 \dimen0=\dimen2\fi%
    \expandafter\edef\csname metricwidth13\endcsname{\the\dimen0}%
    \dimen0=0pt%
    \settowidth{\dimen2}{137.0}\ifdim\dimen2>\dimen0 \dimen0=\dimen2\fi%
    \settowidth{\dimen2}{231.7}\ifdim\dimen2>\dimen0 \dimen0=\dimen2\fi%
    \settowidth{\dimen2}{42.5}\ifdim\dimen2>\dimen0 \dimen0=\dimen2\fi%
    \settowidth{\dimen2}{64.4}\ifdim\dimen2>\dimen0 \dimen0=\dimen2\fi%
    \expandafter\edef\csname metricwidth14\endcsname{\the\dimen0}%
    \dimen0=0pt%
    \settowidth{\dimen2}{137.7}\ifdim\dimen2>\dimen0 \dimen0=\dimen2\fi%
    \settowidth{\dimen2}{271.3}\ifdim\dimen2>\dimen0 \dimen0=\dimen2\fi%
    \settowidth{\dimen2}{43.1}\ifdim\dimen2>\dimen0 \dimen0=\dimen2\fi%
    \settowidth{\dimen2}{65.0}\ifdim\dimen2>\dimen0 \dimen0=\dimen2\fi%
    \expandafter\edef\csname metricwidth15\endcsname{\the\dimen0}%
    \begin{tabular}{l|rrrrr|rrrrr|rrrrr}
      \toprule
       & \multicolumn{5}{c}{BC} & \multicolumn{5}{c}{cDF} & \multicolumn{5}{c}{Ours} \\
      \cmidrule(lr){2-6} \cmidrule(lr){7-11} \cmidrule(lr){12-16}
       & 0.0 & 0.2 & 0.4 & 0.6 & 0.8 & 0.0 & 0.2 & 0.4 & 0.6 & 0.8 & 0.0 & 0.2 & 0.4 & 0.6 & 0.8 \\
      \midrule
      Kubric MOVi-A & 176.1 & 141.2 & 153.8 & 182.6 & 278.7 & 366.2 & 168.5 & 165.8 & 175.5 & 240.0 & 162.1 & 144.6 & 139.2 & {\setlength{\fboxsep}{0pt}\colorbox{blue!18}{\hspace{1.5pt}\makebox[\csname metricwidth14\endcsname][c]{\strut 137.0}\hspace{1.5pt}}} & 137.7 \\
      Kubric MOVi-C & 265.5 & 240.3 & 286.8 & 469.9 & 913.0 & 1353.9 & 376.5 & 345.4 & 454.1 & 894.9 & 319.9 & 250.4 & {\setlength{\fboxsep}{0pt}\colorbox{blue!18}{\hspace{1.5pt}\makebox[\csname metricwidth13\endcsname][c]{\strut 227.0}\hspace{1.5pt}}} & 231.7 & 271.3 \\
      MPMWorlds & 48.2 & 43.9 & 49.0 & 63.3 & 115.5 & 101.1 & 43.2 & 47.3 & 58.8 & 95.5 & 50.6 & {\setlength{\fboxsep}{0pt}\colorbox{blue!18}{\hspace{1.5pt}\makebox[\csname metricwidth12\endcsname][c]{\strut 41.8}\hspace{1.5pt}}} & 42.1 & 42.5 & 43.1 \\
      Rubik's Cube Real & 69.4 & 70.9 & 68.2 & 68.3 & 66.9 & 96.2 & 70.8 & 71.9 & 71.8 & 74.9 & 68.6 & 66.6 & 65.6 & {\setlength{\fboxsep}{0pt}\colorbox{blue!18}{\hspace{1.5pt}\makebox[\csname metricwidth14\endcsname][c]{\strut 64.4}\hspace{1.5pt}}} & 65.0 \\
      \bottomrule
    \end{tabular}%
  }
\end{table}

\begin{table}[H]
  \caption{\textbf{FVD \(\downarrow\) with Wan-5B-LoRA at block size 64.} BC and cDF vary the re-noising used to resist autoregressive drift and Ours varies the transition noise level.}
  \label{tab:context-noise-i3d}
  \centering
  \resizebox{\linewidth}{!}{%
    \dimen0=0pt%
    \settowidth{\dimen2}{241.0}\ifdim\dimen2>\dimen0 \dimen0=\dimen2\fi%
    \settowidth{\dimen2}{264.7}\ifdim\dimen2>\dimen0 \dimen0=\dimen2\fi%
    \settowidth{\dimen2}{35.7}\ifdim\dimen2>\dimen0 \dimen0=\dimen2\fi%
    \settowidth{\dimen2}{105.8}\ifdim\dimen2>\dimen0 \dimen0=\dimen2\fi%
    \expandafter\edef\csname metricwidth1\endcsname{\the\dimen0}%
    \dimen0=0pt%
    \settowidth{\dimen2}{186.1}\ifdim\dimen2>\dimen0 \dimen0=\dimen2\fi%
    \settowidth{\dimen2}{242.6}\ifdim\dimen2>\dimen0 \dimen0=\dimen2\fi%
    \settowidth{\dimen2}{32.0}\ifdim\dimen2>\dimen0 \dimen0=\dimen2\fi%
    \settowidth{\dimen2}{100.3}\ifdim\dimen2>\dimen0 \dimen0=\dimen2\fi%
    \expandafter\edef\csname metricwidth2\endcsname{\the\dimen0}%
    \dimen0=0pt%
    \settowidth{\dimen2}{210.4}\ifdim\dimen2>\dimen0 \dimen0=\dimen2\fi%
    \settowidth{\dimen2}{212.6}\ifdim\dimen2>\dimen0 \dimen0=\dimen2\fi%
    \settowidth{\dimen2}{33.3}\ifdim\dimen2>\dimen0 \dimen0=\dimen2\fi%
    \settowidth{\dimen2}{95.2}\ifdim\dimen2>\dimen0 \dimen0=\dimen2\fi%
    \expandafter\edef\csname metricwidth3\endcsname{\the\dimen0}%
    \dimen0=0pt%
    \settowidth{\dimen2}{202.2}\ifdim\dimen2>\dimen0 \dimen0=\dimen2\fi%
    \settowidth{\dimen2}{236.8}\ifdim\dimen2>\dimen0 \dimen0=\dimen2\fi%
    \settowidth{\dimen2}{40.0}\ifdim\dimen2>\dimen0 \dimen0=\dimen2\fi%
    \settowidth{\dimen2}{91.6}\ifdim\dimen2>\dimen0 \dimen0=\dimen2\fi%
    \expandafter\edef\csname metricwidth4\endcsname{\the\dimen0}%
    \dimen0=0pt%
    \settowidth{\dimen2}{234.0}\ifdim\dimen2>\dimen0 \dimen0=\dimen2\fi%
    \settowidth{\dimen2}{311.1}\ifdim\dimen2>\dimen0 \dimen0=\dimen2\fi%
    \settowidth{\dimen2}{56.1}\ifdim\dimen2>\dimen0 \dimen0=\dimen2\fi%
    \settowidth{\dimen2}{89.6}\ifdim\dimen2>\dimen0 \dimen0=\dimen2\fi%
    \expandafter\edef\csname metricwidth5\endcsname{\the\dimen0}%
    \dimen0=0pt%
    \settowidth{\dimen2}{582.1}\ifdim\dimen2>\dimen0 \dimen0=\dimen2\fi%
    \settowidth{\dimen2}{1115.3}\ifdim\dimen2>\dimen0 \dimen0=\dimen2\fi%
    \settowidth{\dimen2}{141.8}\ifdim\dimen2>\dimen0 \dimen0=\dimen2\fi%
    \settowidth{\dimen2}{121.9}\ifdim\dimen2>\dimen0 \dimen0=\dimen2\fi%
    \expandafter\edef\csname metricwidth6\endcsname{\the\dimen0}%
    \dimen0=0pt%
    \settowidth{\dimen2}{249.3}\ifdim\dimen2>\dimen0 \dimen0=\dimen2\fi%
    \settowidth{\dimen2}{302.7}\ifdim\dimen2>\dimen0 \dimen0=\dimen2\fi%
    \settowidth{\dimen2}{33.2}\ifdim\dimen2>\dimen0 \dimen0=\dimen2\fi%
    \settowidth{\dimen2}{95.5}\ifdim\dimen2>\dimen0 \dimen0=\dimen2\fi%
    \expandafter\edef\csname metricwidth7\endcsname{\the\dimen0}%
    \dimen0=0pt%
    \settowidth{\dimen2}{242.7}\ifdim\dimen2>\dimen0 \dimen0=\dimen2\fi%
    \settowidth{\dimen2}{231.1}\ifdim\dimen2>\dimen0 \dimen0=\dimen2\fi%
    \settowidth{\dimen2}{32.6}\ifdim\dimen2>\dimen0 \dimen0=\dimen2\fi%
    \settowidth{\dimen2}{94.2}\ifdim\dimen2>\dimen0 \dimen0=\dimen2\fi%
    \expandafter\edef\csname metricwidth8\endcsname{\the\dimen0}%
    \dimen0=0pt%
    \settowidth{\dimen2}{266.2}\ifdim\dimen2>\dimen0 \dimen0=\dimen2\fi%
    \settowidth{\dimen2}{255.9}\ifdim\dimen2>\dimen0 \dimen0=\dimen2\fi%
    \settowidth{\dimen2}{37.8}\ifdim\dimen2>\dimen0 \dimen0=\dimen2\fi%
    \settowidth{\dimen2}{90.8}\ifdim\dimen2>\dimen0 \dimen0=\dimen2\fi%
    \expandafter\edef\csname metricwidth9\endcsname{\the\dimen0}%
    \dimen0=0pt%
    \settowidth{\dimen2}{254.7}\ifdim\dimen2>\dimen0 \dimen0=\dimen2\fi%
    \settowidth{\dimen2}{333.2}\ifdim\dimen2>\dimen0 \dimen0=\dimen2\fi%
    \settowidth{\dimen2}{49.9}\ifdim\dimen2>\dimen0 \dimen0=\dimen2\fi%
    \settowidth{\dimen2}{90.4}\ifdim\dimen2>\dimen0 \dimen0=\dimen2\fi%
    \expandafter\edef\csname metricwidth10\endcsname{\the\dimen0}%
    \dimen0=0pt%
    \settowidth{\dimen2}{221.3}\ifdim\dimen2>\dimen0 \dimen0=\dimen2\fi%
    \settowidth{\dimen2}{289.3}\ifdim\dimen2>\dimen0 \dimen0=\dimen2\fi%
    \settowidth{\dimen2}{43.6}\ifdim\dimen2>\dimen0 \dimen0=\dimen2\fi%
    \settowidth{\dimen2}{110.6}\ifdim\dimen2>\dimen0 \dimen0=\dimen2\fi%
    \expandafter\edef\csname metricwidth11\endcsname{\the\dimen0}%
    \dimen0=0pt%
    \settowidth{\dimen2}{176.0}\ifdim\dimen2>\dimen0 \dimen0=\dimen2\fi%
    \settowidth{\dimen2}{228.1}\ifdim\dimen2>\dimen0 \dimen0=\dimen2\fi%
    \settowidth{\dimen2}{32.9}\ifdim\dimen2>\dimen0 \dimen0=\dimen2\fi%
    \settowidth{\dimen2}{102.3}\ifdim\dimen2>\dimen0 \dimen0=\dimen2\fi%
    \expandafter\edef\csname metricwidth12\endcsname{\the\dimen0}%
    \dimen0=0pt%
    \settowidth{\dimen2}{181.0}\ifdim\dimen2>\dimen0 \dimen0=\dimen2\fi%
    \settowidth{\dimen2}{199.0}\ifdim\dimen2>\dimen0 \dimen0=\dimen2\fi%
    \settowidth{\dimen2}{31.1}\ifdim\dimen2>\dimen0 \dimen0=\dimen2\fi%
    \settowidth{\dimen2}{102.8}\ifdim\dimen2>\dimen0 \dimen0=\dimen2\fi%
    \expandafter\edef\csname metricwidth13\endcsname{\the\dimen0}%
    \dimen0=0pt%
    \settowidth{\dimen2}{162.2}\ifdim\dimen2>\dimen0 \dimen0=\dimen2\fi%
    \settowidth{\dimen2}{186.2}\ifdim\dimen2>\dimen0 \dimen0=\dimen2\fi%
    \settowidth{\dimen2}{33.2}\ifdim\dimen2>\dimen0 \dimen0=\dimen2\fi%
    \settowidth{\dimen2}{100.4}\ifdim\dimen2>\dimen0 \dimen0=\dimen2\fi%
    \expandafter\edef\csname metricwidth14\endcsname{\the\dimen0}%
    \dimen0=0pt%
    \settowidth{\dimen2}{171.1}\ifdim\dimen2>\dimen0 \dimen0=\dimen2\fi%
    \settowidth{\dimen2}{214.1}\ifdim\dimen2>\dimen0 \dimen0=\dimen2\fi%
    \settowidth{\dimen2}{33.3}\ifdim\dimen2>\dimen0 \dimen0=\dimen2\fi%
    \settowidth{\dimen2}{99.6}\ifdim\dimen2>\dimen0 \dimen0=\dimen2\fi%
    \expandafter\edef\csname metricwidth15\endcsname{\the\dimen0}%
    \begin{tabular}{l|rrrrr|rrrrr|rrrrr}
      \toprule
       & \multicolumn{5}{c}{BC} & \multicolumn{5}{c}{cDF} & \multicolumn{5}{c}{Ours} \\
      \cmidrule(lr){2-6} \cmidrule(lr){7-11} \cmidrule(lr){12-16}
       & 0.0 & 0.2 & 0.4 & 0.6 & 0.8 & 0.0 & 0.2 & 0.4 & 0.6 & 0.8 & 0.0 & 0.2 & 0.4 & 0.6 & 0.8 \\
      \midrule
      Kubric MOVi-A & 241.0 & 186.1 & 210.4 & 202.2 & 234.0 & 582.1 & 249.3 & 242.7 & 266.2 & 254.7 & 221.3 & 176.0 & 181.0 & {\setlength{\fboxsep}{0pt}\colorbox{blue!18}{\hspace{1.5pt}\makebox[\csname metricwidth14\endcsname][c]{\strut 162.2}\hspace{1.5pt}}} & 171.1 \\
      Kubric MOVi-C & 264.7 & 242.6 & 212.6 & 236.8 & 311.1 & 1115.3 & 302.7 & 231.1 & 255.9 & 333.2 & 289.3 & 228.1 & 199.0 & {\setlength{\fboxsep}{0pt}\colorbox{blue!18}{\hspace{1.5pt}\makebox[\csname metricwidth14\endcsname][c]{\strut 186.2}\hspace{1.5pt}}} & 214.1 \\
      MPMWorlds & 35.7 & 32.0 & 33.3 & 40.0 & 56.1 & 141.8 & 33.2 & 32.6 & 37.8 & 49.9 & 43.6 & 32.9 & {\setlength{\fboxsep}{0pt}\colorbox{blue!18}{\hspace{1.5pt}\makebox[\csname metricwidth13\endcsname][c]{\strut 31.1}\hspace{1.5pt}}} & 33.2 & 33.3 \\
      Rubik's Cube Real & 105.8 & 100.3 & 95.2 & 91.6 & {\setlength{\fboxsep}{0pt}\colorbox{blue!18}{\hspace{1.5pt}\makebox[\csname metricwidth5\endcsname][c]{\strut 89.6}\hspace{1.5pt}}} & 121.9 & 95.5 & 94.2 & 90.8 & 90.4 & 110.6 & 102.3 & 102.8 & 100.4 & 99.6 \\
      \bottomrule
    \end{tabular}%
  }
\end{table}

\subsection{Context noise level}
\label{app:context-noise}

All autoregressive diffusion methods require re-noising of generated outputs to
mitigate autoregressive drift. We sweep the context noise level \(t_c\) for BC
and cDF and the transition noise level \(\tau\) for S2PD from 0.0 to 1.0 in
increments of 0.2, with matched block sizes. Table~\ref{tab:context-noise-cd}
shows that \(t_c=0.2\) works best for BC and cDF and \(\tau=0.6\) works best
for S2PD. S2PD achieves the lowest score across all methods.
\end{samepage}

\subsection{FVD}
\label{app:fvd}

Table~\ref{tab:context-noise-i3d} reports FVD scores, which broadly agree with
the CD-FVD scores. S2PD achieves the lowest FVD on both Kubric datasets and the
MPMWorlds dataset. However, cDF and BC outperform our method on the Rubik's Cube
Real dataset. We attribute this to the limited diversity of the real-world
dataset. For both FVD and CD-FVD, we exclude the condition image and compare
features from temporally subsampled 16-frame windows.

\subsection{MPMWorlds}
\label{app:mpm-metrics}

\inlinepanel[r]{table}{0pt}{%
  \footnotesize
  \setlength{\tabcolsep}{1pt}
  \begin{table}[tbp]
  \centering
  \begin{threeparttable}
  \caption{\textbf{Reimplemented MPMWorlds metrics.} Unofficial metrics implemented based on the paper.}
  \label{tab:mpm-worlds-metrics}
  \dimen0=0pt%
  \settowidth{\dimen2}{0.47}\ifdim\dimen2>\dimen0 \dimen0=\dimen2\fi%
  \settowidth{\dimen2}{0.80}\ifdim\dimen2>\dimen0 \dimen0=\dimen2\fi%
  \settowidth{\dimen2}{0.37}\ifdim\dimen2>\dimen0 \dimen0=\dimen2\fi%
  \settowidth{\dimen2}{0.38}\ifdim\dimen2>\dimen0 \dimen0=\dimen2\fi%
  \settowidth{\dimen2}{0.07}\ifdim\dimen2>\dimen0 \dimen0=\dimen2\fi%
  \settowidth{\dimen2}{0.43}\ifdim\dimen2>\dimen0 \dimen0=\dimen2\fi%
  \expandafter\edef\csname metricwidth1\endcsname{\the\dimen0}%
  \dimen0=0pt%
  \settowidth{\dimen2}{0.45}\ifdim\dimen2>\dimen0 \dimen0=\dimen2\fi%
  \settowidth{\dimen2}{0.72}\ifdim\dimen2>\dimen0 \dimen0=\dimen2\fi%
  \settowidth{\dimen2}{0.34}\ifdim\dimen2>\dimen0 \dimen0=\dimen2\fi%
  \settowidth{\dimen2}{0.31}\ifdim\dimen2>\dimen0 \dimen0=\dimen2\fi%
  \settowidth{\dimen2}{0.10}\ifdim\dimen2>\dimen0 \dimen0=\dimen2\fi%
  \settowidth{\dimen2}{0.40}\ifdim\dimen2>\dimen0 \dimen0=\dimen2\fi%
  \expandafter\edef\csname metricwidth2\endcsname{\the\dimen0}%
  \dimen0=0pt%
  \settowidth{\dimen2}{0.45}\ifdim\dimen2>\dimen0 \dimen0=\dimen2\fi%
  \settowidth{\dimen2}{0.71}\ifdim\dimen2>\dimen0 \dimen0=\dimen2\fi%
  \settowidth{\dimen2}{0.33}\ifdim\dimen2>\dimen0 \dimen0=\dimen2\fi%
  \settowidth{\dimen2}{0.30}\ifdim\dimen2>\dimen0 \dimen0=\dimen2\fi%
  \settowidth{\dimen2}{0.12}\ifdim\dimen2>\dimen0 \dimen0=\dimen2\fi%
  \settowidth{\dimen2}{0.40}\ifdim\dimen2>\dimen0 \dimen0=\dimen2\fi%
  \expandafter\edef\csname metricwidth3\endcsname{\the\dimen0}%
  \dimen0=0pt%
  \settowidth{\dimen2}{0.45}\ifdim\dimen2>\dimen0 \dimen0=\dimen2\fi%
  \settowidth{\dimen2}{0.70}\ifdim\dimen2>\dimen0 \dimen0=\dimen2\fi%
  \settowidth{\dimen2}{0.34}\ifdim\dimen2>\dimen0 \dimen0=\dimen2\fi%
  \settowidth{\dimen2}{0.30}\ifdim\dimen2>\dimen0 \dimen0=\dimen2\fi%
  \settowidth{\dimen2}{0.10}\ifdim\dimen2>\dimen0 \dimen0=\dimen2\fi%
  \settowidth{\dimen2}{0.40}\ifdim\dimen2>\dimen0 \dimen0=\dimen2\fi%
  \expandafter\edef\csname metricwidth4\endcsname{\the\dimen0}%
  \begin{tabular}{lcccc}
    \toprule
     & Parallel & \multicolumn{3}{c}{Serial} \\
    \cmidrule(lr){2-2} \cmidrule(lr){3-5}
    Metric & Bidirectional & cDF & BC & Ours \\
    \midrule
    Foreground mIoU \(\uparrow\) & {\setlength{\fboxsep}{0pt}\colorbox{blue!18}{\hspace{1.5pt}\makebox[\csname metricwidth1\endcsname][c]{\strut 0.47}\hspace{1.5pt}}} & {\setlength{\fboxsep}{0pt}\colorbox{red!18}{\hspace{1.5pt}\makebox[\csname metricwidth2\endcsname][c]{\strut 0.45}\hspace{1.5pt}}} & {\setlength{\fboxsep}{0pt}\colorbox{red!18}{\hspace{1.5pt}\makebox[\csname metricwidth3\endcsname][c]{\strut 0.45}\hspace{1.5pt}}} & {\setlength{\fboxsep}{0pt}\colorbox{red!18}{\hspace{1.5pt}\makebox[\csname metricwidth4\endcsname][c]{\strut 0.45}\hspace{1.5pt}}} \\
    Object collapse \(\downarrow\) & {\setlength{\fboxsep}{0pt}\colorbox{red!18}{\hspace{1.5pt}\makebox[\csname metricwidth1\endcsname][c]{\strut 0.80}\hspace{1.5pt}}} & 0.72 & 0.71 & {\setlength{\fboxsep}{0pt}\colorbox{blue!18}{\hspace{1.5pt}\makebox[\csname metricwidth4\endcsname][c]{\strut 0.70}\hspace{1.5pt}}} \\
    Motion-activity error \(\downarrow\) & {\setlength{\fboxsep}{0pt}\colorbox{red!18}{\hspace{1.5pt}\makebox[\csname metricwidth1\endcsname][c]{\strut 0.37}\hspace{1.5pt}}} & 0.34 & {\setlength{\fboxsep}{0pt}\colorbox{blue!18}{\hspace{1.5pt}\makebox[\csname metricwidth3\endcsname][c]{\strut 0.33}\hspace{1.5pt}}} & 0.34 \\
    Color total variation \(\downarrow\) & {\setlength{\fboxsep}{0pt}\colorbox{red!18}{\hspace{1.5pt}\makebox[\csname metricwidth1\endcsname][c]{\strut 0.38}\hspace{1.5pt}}} & 0.31 & {\setlength{\fboxsep}{0pt}\colorbox{blue!18}{\hspace{1.5pt}\makebox[\csname metricwidth3\endcsname][c]{\strut 0.30}\hspace{1.5pt}}} & {\setlength{\fboxsep}{0pt}\colorbox{blue!18}{\hspace{1.5pt}\makebox[\csname metricwidth4\endcsname][c]{\strut 0.30}\hspace{1.5pt}}} \\
    Temporal anomaly \(\downarrow\) & {\setlength{\fboxsep}{0pt}\colorbox{blue!18}{\hspace{1.5pt}\makebox[\csname metricwidth1\endcsname][c]{\strut 0.07}\hspace{1.5pt}}} & 0.10 & {\setlength{\fboxsep}{0pt}\colorbox{red!18}{\hspace{1.5pt}\makebox[\csname metricwidth3\endcsname][c]{\strut 0.12}\hspace{1.5pt}}} & 0.10 \\
    \cmidrule(lr){1-5}
    Mean composite error \(\downarrow\) & {\setlength{\fboxsep}{0pt}\colorbox{red!18}{\hspace{1.5pt}\makebox[\csname metricwidth1\endcsname][c]{\strut 0.43}\hspace{1.5pt}}} & {\setlength{\fboxsep}{0pt}\colorbox{blue!18}{\hspace{1.5pt}\makebox[\csname metricwidth2\endcsname][c]{\strut 0.40}\hspace{1.5pt}}} & {\setlength{\fboxsep}{0pt}\colorbox{blue!18}{\hspace{1.5pt}\makebox[\csname metricwidth3\endcsname][c]{\strut 0.40}\hspace{1.5pt}}} & {\setlength{\fboxsep}{0pt}\colorbox{blue!18}{\hspace{1.5pt}\makebox[\csname metricwidth4\endcsname][c]{\strut 0.40}\hspace{1.5pt}}} \\
    \bottomrule
  \end{tabular}%
  \end{threeparttable}
\end{table}

}

We obtained the MPMWorlds dataset from its authors~\citep{kovacic2026mpmworlds}.
We retime all video from 30 fps to 24 fps without dropping frames.
Their evaluation code was not available at the time of our experiments,
so we could not run their exact metrics. However, we implemented our own
versions of the metrics based on their paper. Table~\ref{tab:mpm-worlds-metrics} reports these metrics for all four of our methods.
These metrics assess foreground overlap, object collapse, motion, color, and
temporal consistency. Unfortunately, these metrics depend strongly on pixel-wise
alignment with the ground-truth reference video. We used a short text description rather than code conditioning to keep the setup
consistent with the other datasets. Generated samples are not expected to match
their corresponding ground truth videos exactly. This is especially true for
MPMWorlds as physical parameters such as rigidity and viscosity cannot be inferred
from the initial image alone, yet significantly affect how the scene will evolve.
This ambiguity renders these metrics useless to us, but strong code conditioning
is an interesting direction for future work.

\section{Additional Qualitative Results}
\label{app:qualitative}

\begin{figure}[!b]
  \centering
  \includegraphics[width=0.97\linewidth]{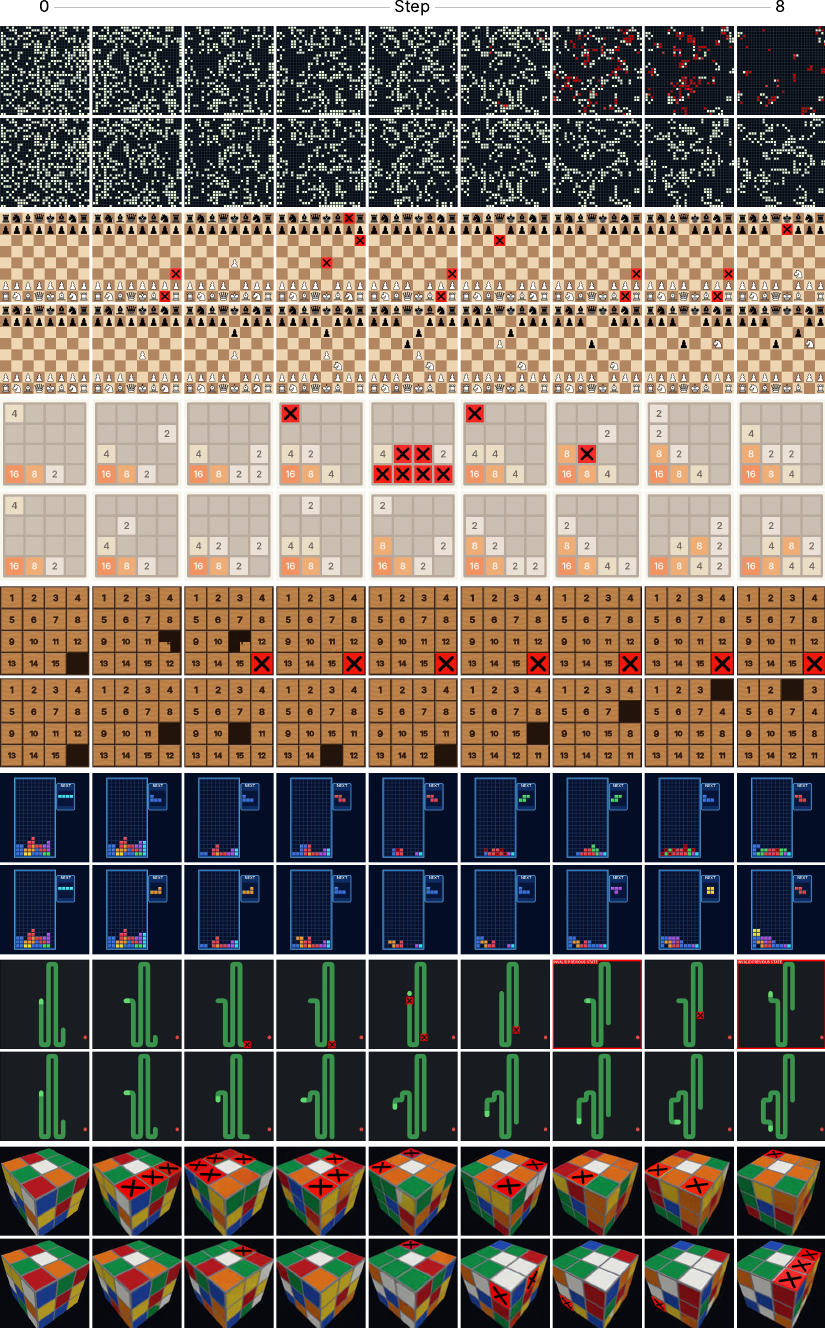}
  \setlength{\abovecaptionskip}{3pt}
  \setlength{\belowcaptionskip}{0pt}
  \small
  \caption{\textbf{Discrete datasets.} Qualitative comparison on Conway, Chess, 2048, Fifteen Puzzle, Tetris, Snake, and Rubik's Cube 3D. For each dataset, the top row shows Bidirectional and the bottom row shows Ours. We show select frames from the ten-second video samples.}
  \label{fig:qualitative-discrete}
\end{figure}

\begin{figure}[tbp]
  \centering
  \includegraphics[width=0.97\linewidth]{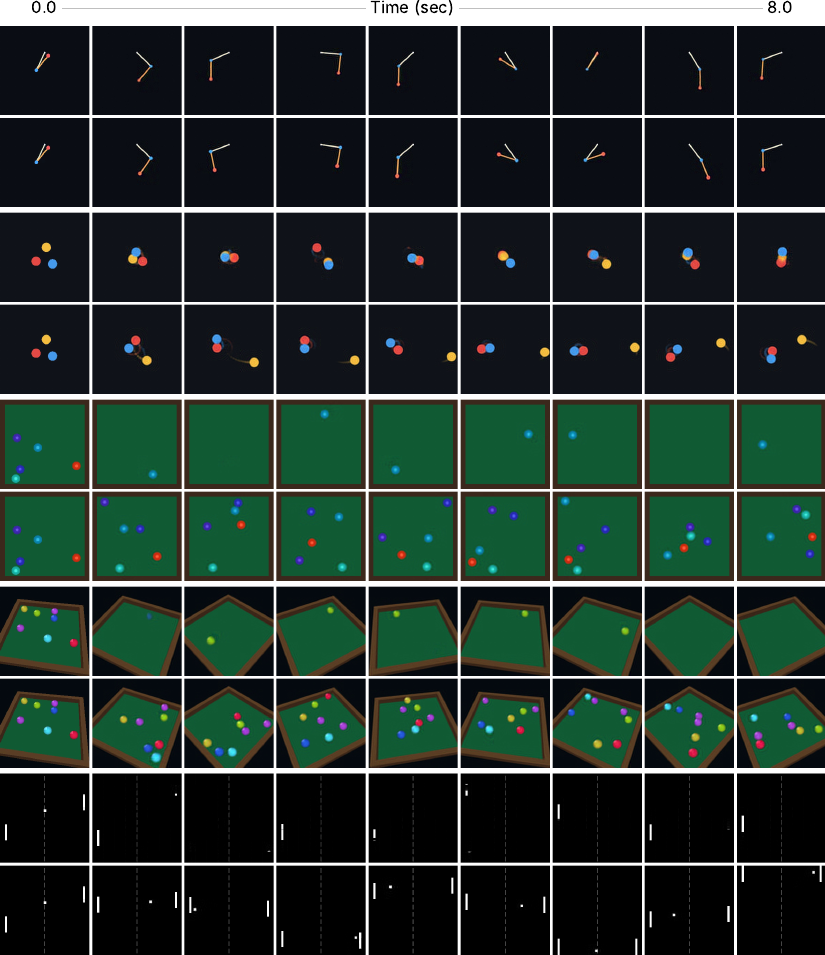}
  \setlength{\abovecaptionskip}{3pt}
  \setlength{\belowcaptionskip}{0pt}
  \small
  \caption{\textbf{Continuous datasets.} Qualitative comparison on Double Pendulum, Three Body, Colliding Balls 2D, Colliding Balls 3D, and Pong. For each dataset, the top row shows Bidirectional and the bottom row shows Ours. We show select frames from the ten-second video samples.}
  \label{fig:qualitative-continuous}
\end{figure}

\begin{figure}[tbp]
  \centering
  \includegraphics[width=0.97\linewidth]{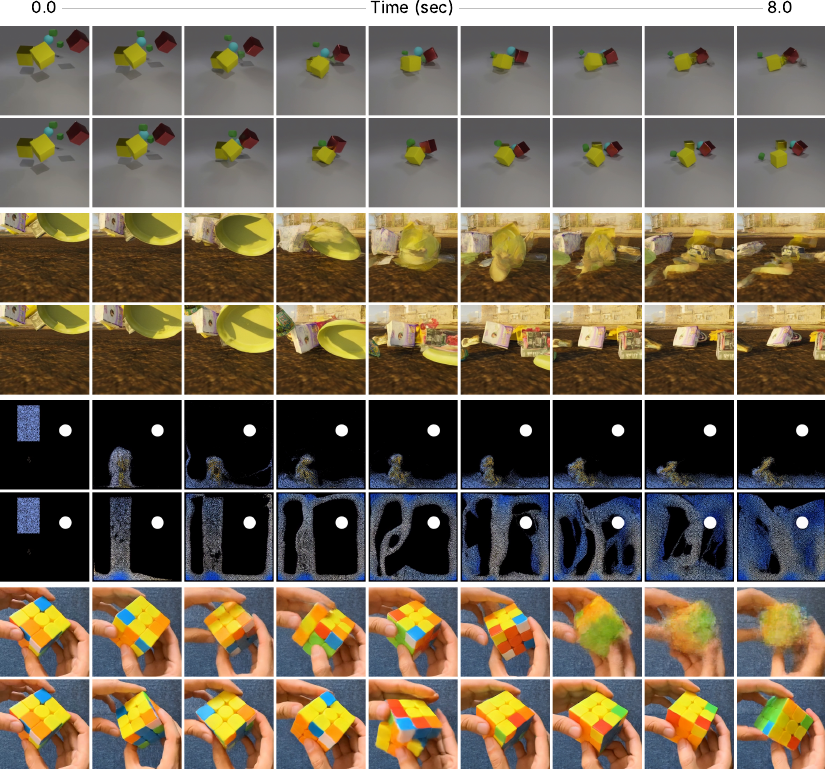}
  \setlength{\abovecaptionskip}{3pt}
  \setlength{\belowcaptionskip}{0pt}
  \small
  \caption{\textbf{Video datasets.} Qualitative comparison on Kubric MOVi-A, Kubric MOVi-C, MPMWorlds, and Rubik's Cube Real. For each dataset, the top row shows Bidirectional and the bottom row shows Ours. We show select frames from the ten-second video samples.}
  \label{fig:qualitative-video}
\end{figure}

Here we show frames from videos generated by our method and the bidirectional
baseline for each dataset. In Figure~\ref{fig:qualitative-discrete},
the x axis shows step count for each game. Each step spans four frames at
24 frames per second, so steps 0--8 show selected frames from the first
1.33 seconds of each 10-second sample. We exclude intermediate animation
frames and show consecutive settled states---the same states parsed for
evaluation. Figures~\ref{fig:qualitative-continuous}
and~\ref{fig:qualitative-video} show one frame per second for the
first eight seconds of each 10-second sample. The first frame of each row
is the conditioning image.

\end{document}